\pdfoutput=1

\documentclass[11pt]{article}

\usepackage[final]{acl}

\usepackage{times}
\usepackage{latexsym}

\usepackage[T1]{fontenc}

\usepackage[utf8]{inputenc}

\usepackage{microtype}

\usepackage{inconsolata}

\usepackage{graphicx}

\usepackage{booktabs}
\usepackage{multirow}
\usepackage{amsmath}
\usepackage{arydshln}
\usepackage{adjustbox}
\usepackage{color,xcolor,colortbl}
\usepackage[framemethod=TikZ]{mdframed}
\usepackage{pgfplots}
\usepgfplotslibrary{fillbetween}
\usetikzlibrary{patterns}
\usepackage{subcaption}

\pgfplotsset{compat=1.18}

\makeatletter
\def\adl@drawiv#1#2#3{%
        \hskip.5\tabcolsep
        \xleaders#3{#2.5\@tempdimb #1{1}#2.5\@tempdimb}%
                #2\z@ plus1fil minus1fil\relax
        \hskip.5\tabcolsep}
\newcommand{\cdashlinelr}[1]{%
  \noalign{\vskip\aboverulesep
           \global\let\@dashdrawstore\adl@draw
           \global\let\adl@draw\adl@drawiv}
  \cdashline{#1}
  \noalign{\global\let\adl@draw\@dashdrawstore
           \vskip\belowrulesep}}
\makeatother

\newcommand{\quotes}[1]{``#1''}
\renewcommand{\vec}[1]{\mathbf{#1}}

\definecolor{lightblue}{HTML}{5DA5DA}
\definecolor{lightorange}{HTML}{FAA43A}
\definecolor{lightgreen}{HTML}{60BD68}
\definecolor{lightpurple}{HTML}{B276B2}

\title{Generative Models Enhanced by Sequence Labelling and Aspect-Code Switching Improve Cross-lingual Aspect-Based Sentiment Analysis}

\author{
 Jakub \v{S}m\'{i}d\textsuperscript{*, $\dagger$} \and
 Pavel P\v{r}ib\'{a}\v{n}\textsuperscript{*} \and
 Pavel Kr\'{a}l\textsuperscript{*, $\dagger$}
 \vspace{8pt}
\\
 \textsuperscript{*}Department of Computer Science and Engineering\\
 \textsuperscript{$\dagger$}NTIS -- New Technologies for the Information Society\\
 University of West Bohemia in Pilsen, Faculty of Applied Sciences \\
         Univerzitní 2732/8, 301 00 Pilsen, Czech Republic \\
    \texttt{\{jaksmid,pribanp,pkral\}@fav.zcu.cz}\\
         \url{https://nlp.kiv.zcu.cz}
}

\begin{document}
\maketitle

\begin{abstract}

Cross-lingual aspect-based sentiment analysis (ABSA) transfers knowledge from a source language with annotated data to a target language, enabling fine-grained sentiment analysis without annotated target-language data. While monolingual ABSA has seen significant progress, cross-lingual ABSA remains underexplored, especially for complex tasks involving multiple sentiment elements like target-aspect-sentiment detection (TASD). In this paper, we propose a novel SeqLab framework that enhances cross-lingual ABSA using a sequence-to-sequence model with an auxiliary sequence-labelling task performed by the encoder, enhancing aspect term recognition and sentiment predictions. Additionally, we incorporate aspect-code switching (ACS), a translation-based technique that swaps aspect terms between source and translated sentences, generating additional training data to enhance the model’s cross-lingual understanding. We evaluate our approach across eleven languages, three domains, and two backbone models, surpassing previous state-of-the-art results for the commonly studied E2E-ABSA task. Unlike most prior work that relies solely on English as the source language, we systematically assess different source-target language pairs and extend our evaluation to the more challenging, yet underexplored TASD task in cross-lingual settings. Finally, we provide a detailed error analysis highlighting key challenges and limitations.

\end{abstract}

\section{Introduction}

Aspect-based sentiment analysis (ABSA) identifies sentiment towards specific aspects within a text, making it essential for evaluating products, services, and user experiences~\citep{zhang2022survey}. It involves extracting sentiment elements such as aspect terms, aspect categories, and sentiment polarities. For example, in \textit{\quotes{Tasty apple}}, \textit{\quotes{apple}} is the aspect term, categorized under \textit{\quotes{food quality}} with a \textit{\quotes{positive}} sentiment. Despite its broad applicability, most ABSA research focuses on English due to the high cost of annotation. This work addresses zero-shot cross-lingual ABSA, aiming to transfer knowledge from a labelled source language to target languages without annotated data.

Traditional cross-lingual approaches rely on translation-based annotation~\citep{lambert-2015-aspect, zhou2015clopinionminer} or cross-lingual embeddings~\citep{barnes-etal-2016-exploring, akhtar-etal-2018-solving}, while recent methods~\citep{zhang-etal-2021-cross, LinXABSA} leverage multilingual pre-trained language models (mPLMs) such as mBERT~\citep{devlin-etal-2019-bert} and XLM-R~\citep{conneau-etal-2020-unsupervised}. However, zero-shot cross-lingual ABSA remains challenging due to language-specific aspect terms, informal expressions, and the limited representation of low-resource languages in mPLMs~\citep{li2020unsupervised}.

Most cross-lingual studies focus on simpler tasks involving individual sentiment elements or E2E-ABSA, where aspect terms and their sentiment polarity are predicted jointly. In contrast, monolingual research has shifted towards sequence-to-sequence models and more complex tasks, such as target-aspect-sentiment detection (TASD)~\citep{tasd}, which simultaneously extracts aspect terms, aspect categories, and sentiment polarity. Despite these advancements, complex tasks and sequence-to-sequence models remain largely unexplored in cross-lingual ABSA~\citep{SMID2025103073}.

To address the limitations of prior cross-lingual ABSA research, we introduce a novel \textsc{ACS + SeqLab} framework that combines sequence-to-sequence modelling with auxiliary sequence labelling and translation-based augmentation. The \textsc{SeqLab} component enhances aspect term recognition by leveraging encoder-level sequence-labelling signals both during training and for refining generated sentiment tuples at inference. In parallel, aspect-code switching (\textsc{ACS}) augments the data by swapping aspect terms between source and machine-translated sentences, enabling the model to better capture language-specific variations. These components complement each other, with \textsc{ACS} providing target-language data augmentation and \textsc{SeqLab} encouraging the generative backbone to maintain greater awareness of word-level boundaries. Together, these components substantially improve cross-lingual ABSA performance.

Our key contributions are: 1) We propose a novel framework that combines an alignment-free translation method, aspect-code switching, and generative models enhanced with sequence labelling to improve cross-lingual ABSA. 2) We evaluate our method across eleven languages, three domains, and two backbone models in various source-target language combinations, achieving new state-of-the-art results. 3) We extend our framework to the underexplored and challenging cross-lingual TASD task. 4) We conduct a detailed analysis to identify key challenges and limitations of the proposed framework, offering insights for future research.

\section{Related Work}
\label{sec:related}

Early cross-lingual ABSA research primarily focused on aspect term extraction and sentiment classification, relying on translation-based annotation and word alignment tools like FastAlign~\citep{dyer-etal-2013-simple} to obtain pseudo-labelled data~\citep{zhou2015clopinionminer, lambert-2015-aspect} or cross-lingual word embeddings to enable language-agnostic ABSA~\citep{wang2018transition, barnes-etal-2016-exploring}.

More recent approaches have addressed mainly the E2E-ABSA task using multilingual encoder-only Transformer models, such as mBERT~\citep{devlin-etal-2019-bert} and XLM-R~\citep{conneau-etal-2020-unsupervised}, often combined with machine translation. \textsc{Translation-TA}~\citep{li2020unsupervised} follows the \textit{Translate-then-Align} paradigm, fine-tuning on translated data, while \textsc{Bilingual-TA} extends this by incorporating both translated and original source data. \textsc{ACS}~\citep{zhang-etal-2021-cross} introduces an alignment-free projection method with aspect code-switching, and \textsc{ACS-Distill} enhances it with distillation on unlabelled target-language data. \textsc{CL-XABSA}~\citep{lin2023clxabsa} integrates contrastive learning, while \textsc{Equi-XABSA}~\citep{LinXABSA} employs a dynamically weighted loss to address class imbalances and leverages anti-decoupling for improved semantic representation.

Some methods avoid machine translation and are not limited to encoder-only models. \textsc{LACA}~\citep{smid-etal-2025-laca} leverages large language models (LLMs) to generate pseudo-labeled targets using various backbone architectures. \citet{wu2025evaluating} study LLMs across multiple settings, including zero-shot and few-shot prompting, as well as zero-shot cross-lingual fine-tuning. \citet{wu-etal-2025-absa} employ mT5 and, in addition to E2E-ABSA, also address the TASD task, as do \citet{icaart25}, who improve cross-lingual performance through constrained decoding (CD).

\section{Methodology}
\begin{figure}[ht!]
    \centering
    \includegraphics[width=0.8\linewidth, trim=8pt 0pt 0pt 0pt, clip]{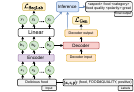}
    \caption{Overview of the proposed framework. The encoder's auxiliary sequence labelling task improves the model and final output.  The decoder input consists of the gold tokens from the label, and the output is the model's prediction.}
    \label{fig:overview}
\end{figure}

This section describes our framework for the TASD task, applicable to the E2E-ABSA task by omitting aspect categories. Figure~\ref{fig:overview} shows the framework overview, which leverages a sequence-to-sequence model, auxiliary sequence labelling task, and combined inference to improve the final output.

\subsection{Problem Formulation}
Given an input sentence $\boldsymbol{x} = \{x_i\}_{i=1}^n$ containing $n$ tokens, the aim is to predict all sentiment tuples $\boldsymbol{y}=\{t_i\}_{i=1}^{|t|}$, where each tuple $t_i$ comprises an aspect term ($a$), aspect category ($c$), and sentiment polarity ($p$). Alternatively, ABSA can be modelled as a sequence labelling problem, where the task is to predict token-level labels for the input sentence.

In cross-lingual ABSA, the goal is to predict labels $\boldsymbol{y}^\mathcal{T}$ for sentences $\boldsymbol{x}^\mathcal{T}$ in the target language $\mathcal{T}$ using only labelled data $(\boldsymbol{x}^\mathcal{S}, \boldsymbol{y}^\mathcal{S})$ from the source language $\mathcal{S}$. The dataset $\mathcal{D}_\mathcal{S}$ does not include labelled examples in the target language, making this a zero-shot cross-lingual task.

\subsection{ABSA Generative Model Assisted with Sequence Labelling}
\label{sub:model}

Figure~\ref{fig:overview} illustrates the proposed framework, which uses a sequence-to-sequence (encoder–decoder) model to generate the output, assisted by an auxiliary sequence labelling task. We initialize the ABSA system with weights from pre-trained multilingual models that form the backbone of the architecture. During training, we fine-tune the full set of parameters $\vec{\Theta}$, including task-specific weights $\vec{W}$ and biases $\vec{b}$.

\subsubsection{Generative Framework}

We utilize sequence-to-sequence models, where the encoder processes the input sequence $\boldsymbol{x}$ into a contextualized representation, and the decoder generates the output sequence $\boldsymbol{y}$ token by token. Each token $y_i$ is predicted based on prior tokens $y_{1}^{i-1}$ and the encoded representation.

To enhance the generative process, we introduce special tokens: \texttt{<aspect>}, \texttt{<category>}, \texttt{<polarity>}, \texttt{<ssep>} (tuple separator), and \texttt{<null>} (for sentences with no sentiment tuples). These tokens are added to the tokenizer and assigned new embeddings. Aspect categories, originally in the format \quotes{ENTITY\#ATTRIBUTE}, are converted into natural language (e.g. \quotes{FOOD\#QUALITY} becomes \textit{\quotes{food quality}}). We map sentiment polarities: \textit{\quotes{positive}} to \textit{\quotes{great}}, \textit{\quotes{negative}} to \textit{\quotes{bad}}, and \textit{\quotes{neutral}} to \textit{\quotes{ok}}. Figure~\ref{fig:overview} illustrates an example of the output structure.

We fine-tune the model by minimizing the cross-entropy loss $\mathcal{L}_{\text{Gen}}$ for the generated sequence:

\begin{equation}
    \small
    \mathcal{L\textsubscript{Gen}} = \frac{1}{|\mathcal{D}|}\sum_{(\boldsymbol{x},\boldsymbol{y})\in \mathcal{D}}\left[-\frac{1}{n}\sum_{i=1}^n\log P_{\vec{\Theta}}(y_i|\boldsymbol{x},y_1^{i-1})\right].
\end{equation}

\subsubsection{Sequence Labelling Assistance}

Inspired by prior work in monolingual ABSA~\citep{xianlong-etal-2023-tagging}, we enhance the generative process with sequence labelling by leveraging the encoder's output. The encoder produces hidden vectors $\vec{h} = \{\vec{h}_i\}_{i=1}^n$ for each token in the input $\boldsymbol{x}$. Using a tagging scheme, we assign token-level labels $\mathcal{Y} = \{\mathtt{T\text{-}POS, T\text{-}NEG, T\text{-}NEU, O}\}$, indicating whether a token is part of an aspect term, labelling only the first token of each word. For example, $\mathtt{T\text{-}POS}$ means the token is part of a positive aspect term. Finally, a linear classification layer maps the hidden vectors $\vec{h}_i$ to label probabilities:

\begin{equation}
    P_\vec{\Theta}(y_i|x_i)={\operatorname{softmax}}(\vec{W}\vec{h}_i + \vec{b}).
\end{equation}

This scheme does not fully capture multi-word aspect terms, which would require a BIO or similar tagging approach commonly used in encoder-only models. It also does not encode information about aspect categories. However, this is not necessary in our approach, as sequence labelling serves only as an auxiliary task. We keep the process simple and efficient by focusing on individual words from the encoder output rather than entire aspect phrases.

The cross-entropy loss for sequence labelling, $\mathcal{L}_{\text{SeqLab}}$, is defined as:
\begin{equation}
    \small
    \mathcal{L\textsubscript{SeqLab}} = \frac{1}{|\mathcal{D}|}\sum_{(\boldsymbol{x},\boldsymbol{y})\in \mathcal{D}}\left[-\frac{1}{n}\sum_{i=1}^ny_i\log P_\vec{\Theta}(y_i|x_i)\right].
\end{equation}

\subsubsection{Joint Optimization}
The final loss $\mathcal{L}$ combines the generative and sequence labelling losses:
\begin{equation}
    \mathcal{L} = \mathcal{L\textsubscript{Gen}} + \alpha \cdot \mathcal{L\textsubscript{SeqLab}},
\end{equation}
where $\alpha$ is a hyperparameter controlling the weight of the auxiliary sequence labelling task.

\subsubsection{Inference}

\begin{figure}[ht!]
    \centering
    \includegraphics[width=0.65\linewidth]{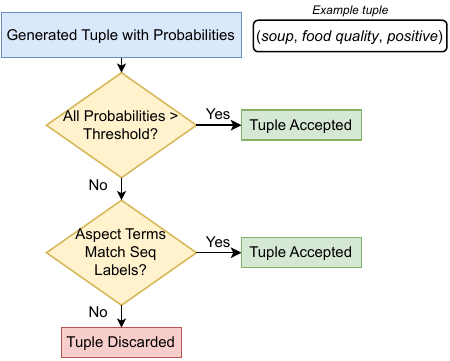}
    \caption{Inference algorithm.}
    \label{fig:inference}
\end{figure}

\begin{figure*}
    \centering
    \includegraphics[width=\linewidth]{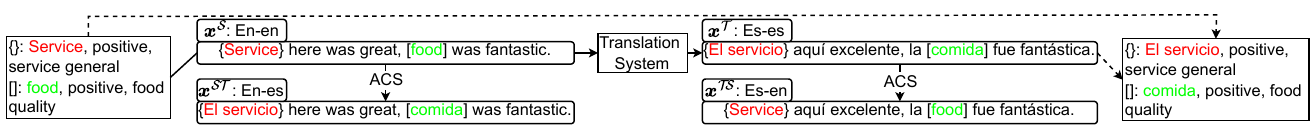}
    \caption{Example of label projection method with aspect code-switching (lower part) for English–Spanish pair.}
    \label{fig:acs}
\end{figure*}

Similar to \citet{xianlong-etal-2023-tagging}, we leverage the sequence labelling task output to enhance the generated outputs' quality. Specifically, we first extract the probabilities associated with each word in the tuples generated by the model. For each tuple, if all the words it contains have probabilities exceeding an experimentally determined threshold, the tuple is retained without further checks. However, if any word in the tuple falls below this probability threshold, we compare the generated tuple with the aspect terms identified during the sequence labelling task performed by the encoder. In this comparison, if the aspect term in the generated tuple is a subset of the aspect terms identified by sequence labelling, or vice versa, the tuple is retained. Otherwise, we discard the tuple. This approach, depicted in Figure~\ref{fig:inference}, ensures that the generated outputs are consistent with the aspect terms identified by the sequence labelling task while filtering out low-confidence or inconsistent tuples.

\subsection{Aspect Code-Switching Mechanism}
\label{sec:acs}

\citet{zhang-etal-2021-cross} demonstrate that alignment-free translation and aspect code-switching (\textsc{ACS}) – swapping aspect terms between source and target languages – improves cross-lingual ABSA. \citet{smid-etal-2024-uwb} also applied this idea to the cross-lingual detection of emotion-triggering words. Therefore, we adopt it too.

We begin by marking aspect terms in the source sentence with special symbols (e.g. \quotes{\{\}} or \quotes{[]}) to ensure they remain identifiable after translation. We use unique symbols to prevent ambiguity for sentences with multiple aspects. We then translate the sentence and retrieve aspect terms using these markers, as illustrated in Figure~\ref{fig:acs}.

Next, given a source sentence $\boldsymbol{x}^\mathcal{S}$ and its translation $\boldsymbol{x}^\mathcal{T}$, we generate two additional aspect code-switched sentences: $\boldsymbol{x}^\mathcal{ST}$, where the source sentence contains aspect terms in the target language, and $\boldsymbol{x}^\mathcal{TS}$, where the translated sentence contains the aspect terms in the source language, as shown in the lower part of Figure~\ref{fig:acs}. The final \textsc{ACS} dataset comprises four subsets: source dataset $\mathcal{D}_\mathcal{S}$, translated dataset $\mathcal{D}_\mathcal{T}$, and aspect-code switched datasets $\mathcal{D}_\mathcal{ST}$ and $\mathcal{D}_\mathcal{TS}$. During training, we merge the four subsets into a single pool and sample instances uniformly at random.

\section{Experimental Setup}
We conduct experiments on E2E-ABSA and TASD.

\subsection{Dataset}
We evaluate the proposed framework on the SemEval-2016 dataset~\citep{pontiki-etal-2016-semeval}, which includes real user restaurant reviews in English (en), Spanish (es), French (fr), Dutch (nl), Russian (ru), and Turkish (tr). We use the data splits provided by \citet{zhang-etal-2021-cross} for a fair comparison. Additionally, we use the hotel and laptop domain pre-split datasets from \citet{wu-etal-2025-absa}. Unlike the restaurant dataset, this dataset is parallel; only the English portion contains original reviews, while the other languages are translations. We consider the same six languages as above and further include Japanese (ja), Korean (ko), Thai (th), Vietnamese (vi), and Chinese (zh) to broaden the linguistic coverage. Appendix~\ref{app:data_stats} presents the datasets statistics. We use the source language validation set for model selection in all experiments to ensure true unsupervised settings~\citep{zhang-etal-2021-cross}.

\subsection{Implementation Details}
We employ the base and large mT5 models~\citep{xue-etal-2021-mt5} as well as the large mBART model~\cite{tang2020multilingual} in our experiments. 
Translations are obtained using the Google Translate API\footnote{\url{https://translate.google.com/}} as described in Section~\ref{sec:acs}. Appendix~\ref{app:details} provides the full experimental setup.

\subsection{Evaluation Metrics}
We employ micro-F1 as the evaluation metric, consistent with related work~\citep{zhang-etal-2021-cross, lin2023clxabsa, LinXABSA}, where we deem a predicted tuple correct only if all sentiment elements within the tuple are accurate. We report average F1 scores across five runs with different random seeds.

\subsection{Compared Methods}

We compare our \textsc{SeqLab} framework against several methods. The \textsc{Zero-shot} baseline fine-tunes the model using only labelled source-language data, a strong benchmark for cross-lingual tasks~\citep{conneau-etal-2020-unsupervised, wu-dredze-2019-beto}. The remaining methods are described in Section~\ref{sec:related}. Most prior work focuses on the restaurant domain and the E2E-ABSA task, typically using encoder-only models like XLM-R. These models differ in their pre-training data to mT5 we use, which may affect overall and language-specific performance. Some studies do not include Turkish as a target language due to its relatively small dataset size.

Most compared methods share the same data and task formulation. However, \citet{wu-etal-2025-absa} and \citet{icaart25} also predict implicit aspect terms (e.g. in sentence \textit{\quotes{Delicious}}), making their task definition differ from ours. We exclude implicit aspect terms -- consistent with the majority of prior work -- to ensure a uniform setup and enable direct comparison. We also include a comparison with GPT-4o mini~\citep{openai2024gpt4ocard}, using the prompt provided in Appendix~\ref{app:details}.

\section{Results}

Key observations of the restaurant domain results shown in Table~\ref{tab:res} include: 
\\
\hspace*{5pt} 
1) The mT5 model consistently outperforms encoder-based models in the \textsc{Monolingual} settings across all target languages, with an average improvement of over 4.5\%. The only exception is French, where the performance is comparable. This disparity is likely due to differences in pre-training corpora between the models. Incorporating the \textsc{SeqLab} method further enhances performance by nearly 2\% on average.
\begin{table}[ht!]
    \centering
    \begin{adjustbox}{width=\linewidth}
    
        \begin{tabular}{@{}lllcccccc@{}}
        \toprule
       \textbf{Task} & \textbf{Method}     & \textbf{Es}    & \textbf{Fr}    & \textbf{Nl}    & \textbf{Ru} & \textbf{Tr}    & \textbf{Avg\textsuperscript{\S}}   \\ \midrule
       \multirow{20}{*}{\rotatebox[origin=c]{90}{E2E-ABSA}} & GPT-4o mini$^\dagger$ & 50.02 & 48.21 & 49.14 & 45.19 & 45.78 & 48.14 \\ 
       & GPT-4o {\footnotesize \citep{wu2025evaluating}}$^\dagger$ & 54.01 & 52.28 & 53.06 & 44.88 & -- & 51.06 \\
       &\textsc{Monolingual} {\footnotesize \citep{zhang-etal-2021-cross}}*       & 71.93          & 67.44          & 64.28          & 64.93     & --     & 67.15          \\
        &        \textsc{Monolingual}               & 75.18          & 67.80          & 71.39          & 72.52  & 54.98        & 71.72          \\
        &\phantom{0} + \textsc{SeqLab}      & 77.18          & 70.35          & 73.84          & 73.23  & 62.55         & 73.65 \\ \cdashlinelr{2-8}
        &\textsc{Zero-shot} {\footnotesize \citep{li2020unsupervised}}*           & 67.10          & 56.43          & 59.03          & 56.80    &  46.21     & 59.84          \\
        &\textsc{Translation-TA} {\footnotesize \citep{li2020unsupervised}}*          & 58.10          & 47.00          & 56.19          & 50.34    & 40.24      & 52.91          \\
        &\textsc{Bilingual-TA} {\footnotesize \citep{li2020unsupervised}}*             & 61.87          & 49.34          & 58.64          & 52.89  & 41.44        & 55.69          \\
        &\textsc{ACS} {\footnotesize \citep{zhang-etal-2021-cross}}*         & 67.32          & 59.39          & 62.83          & 60.81      & --    & 62.59          \\
        &\textsc{ACS-Distill} {\footnotesize \citep{zhang-etal-2021-cross}}* & 69.24          & 59.90          & 63.74          & 62.02    & --      & 63.73          \\
        &\textsc{CL-XABSA} {\footnotesize \citep{lin2023clxabsa}}*   & 64.85          & 59.47          & 59.75          & 61.13  & --        & 61.16          \\
        & Fine-tuned LLMs {\footnotesize \citep{wu2025evaluating}}$^\dagger$ & 68.95 & 63.01 & 60.84 & 53.50 & -- & 61.58 \\
        &\textsc{Equi-XABSA} {\footnotesize \citep{LinXABSA}}*           & 69.56          & 60.68          & 61.31          & 62.34        & --  & 63.47 
         \\
         & \textsc{mT5-large-CD} {\footnotesize \citep{icaart25}}\textsuperscript{$\dagger,\ddagger$} & 69.30 & 61.10 & 60.80 & 63.70 & 48.90 & 63.73 \\
         &\textsc{mT5-LACA} {\footnotesize \citep{smid-etal-2025-laca}} & 72.03 & \textbf{63.92} & 65.70 & 62.92 & -- & 65.92 \\
         &\textsc{LLM-LACA} {\footnotesize \citep{smid-etal-2025-laca}}$^\dagger$ & \underline{74.27} & \underline{70.13} & 68.25 & 62.38 & -- & 68.76 \\
         \cdashlinelr{2-8}
        &\textsc{Zero-shot}                & 62.30          & 57.83          & 66.46          & 60.64     & 46.23     & 61.81          \\
        &\phantom{0} + \textsc{SeqLab}       & 63.35          & 60.18          & 69.07          & 63.36     & 48.91     & 63.99          \\
        &\textsc{ACS}                 & 70.84          & 62.30          & 71.52          & 67.96      & 52.92    & 68.16          \\
        &\phantom{0} + \textsc{SeqLab}        & \textbf{72.56} & 62.42 & \textbf{72.00} & \textbf{\underline{68.99}} & \textbf{\underline{53.02}} & \textbf{\underline{68.99}} \\ \midrule
        \multirow{9}{*}{\rotatebox[origin=c]{90}{TASD}} & GPT-4o mini$^\dagger$ & 40.92 & 39.80 & 35.13 & 32.66 & 32.41 & 37.13 \\
        & GPT-4o {\footnotesize \citep{wu2025evaluating}}$^\dagger$ & 48.89 & 46.76 & 42.14 & 37.65 & -- & 45.36 \\
        & \textsc{Monolingual} & 71.20 & 58.93 & 64.35 & 69.57 & 36.84 & 66.01\\
        &\phantom{0} + \textsc{SeqLab} & 72.36 & 60.35 & 67.39 & 70.30 & 49.88 & 67.60 \\ \cdashlinelr{2-8}
        & \textsc{mT5-large-CD} {\footnotesize \citep{icaart25}}\textsuperscript{$\dagger,\ddagger$} & 57.60 & 50.40 & 50.40 & 54.90 & 43.80 & 53.33 \\ \cdashlinelr{2-8}
        & \textsc{Zero-shot} & 54.60 & 49.81 & 59.64 & 55.08 & 39.61 & 54.78 \\
        & \phantom{0} + \textsc{SeqLab} & 58.00 & 51.79 & 62.25 & 58.05 & 43.42 & 57.52 \\ 
        & \textsc{ACS} & 64.98 & 55.14 & 65.79 & 63.55 & \textbf{49.41} & 62.37 \\
        & \phantom{0} + \textsc{SeqLab} & \textbf{66.28} & \textbf{56.30} & \textbf{66.58} & \textbf{64.27} & 46.35 & \textbf{63.36} \\
        \bottomrule
        \end{tabular}
    \end{adjustbox}
    \caption{Results on the restaurant domain with English as the source language and other languages as targets. The best result in the cross-lingual setting for each language and task for base-sized models is highlighted in \textbf{bold}, for models not taking into account size \underline{underlined}. \textsuperscript{\S} denotes that average scores excludes Turkish. * indicates prior works using encoder-only models (XLM-R; mBERT in \citet{lin2023clxabsa}). $^\dagger$ indicates methods with larger models. $^\ddagger$ denotes different task formulation.}
    \label{tab:res}
\end{table}
\\
\hspace*{5pt} 
2) In \textsc{Zero-shot} settings, mT5 surpasses XLM-R by almost 2\% on average, with notable gains for Dutch (+7.43\%) and Russian (+3.84\%), but a drop in Spanish by nearly 5\%. The \textsc{SeqLab} method improves average zero-shot results by over 2\%.
\\
\hspace*{5pt} 
3) Incorporating \textsc{ACS} data into fine-tuning significantly boosts performance, improving results by over 7\% and 4\% compared to \textsc{Zero-shot} and \textsc{Zero-shot + SeqLab}, respectively. This highlights the importance of target-language data, even when pseudo-labelled, in bridging the source–target gap. Adding \textsc{SeqLab} further improves performance by approximately 1\% over \textsc{ACS} alone. 
\\
\hspace*{5pt} 
4) Our \textsc{ACS + SeqLab} approach outperforms the previous best base-sized model (\textsc{mT5-LACA}) by over 3\% on average. While performance is slightly lower for French and comparable for Spanish, it yields substantial gains for Dutch and Russian, highlighting the continued effectiveness of translation-based approaches compared to LLM-based augmentation. Compared to prior translation-based methods such as \textsc{ACS-Distill}, our method improves results by more than 5\%.
\\
\hspace*{5pt} 
5) The proposed \textsc{ACS + SeqLab} framework significantly outperforms most LLM baselines. Compared to the best GPT-4o results, it achieves nearly 18\% higher performance in cross-lingual settings and exceeds fine-tuned LLMs by more than 7\% on average. At the same time, our approach achieves comparable average performance to \textsc{LLM-LACA}, which combines LLM fine-tuning with LLM-based pseudo-labelling. While \textsc{LLM-LACA} performs better on Spanish and French, our method is stronger on Dutch and Russian, likely due to differences in language support in the underlying \textsc{LLaMA 3.1} models~\citep{dubey2024llama3herdmodels}. Importantly, our approach is substantially more efficient, relying on a 580M-parameter model compared to 8B-70B models used in \textsc{LLM-LACA}, resulting in significantly lower computational and memory requirements. Moreover, data translation is generally less resource-intensive than LLM-based data generation. Overall, both approaches achieve strong cross-lingual performance, but the choice depends on computational constraints and target language. 
\\
\hspace*{5pt}
6) Among the target languages, French yields the lowest overall performance, while Dutch and Spanish exhibit the highest scores. This could be attributed to Dutch's linguistic similarity to English and the higher presence of Spanish in mT5's pre-training data.
\\
\hspace*{5pt}
7) The best overall results for each language are achieved by combining \textsc{SeqLab} and \textsc{ACS}. On average, this combination surpasses the previous best method by over 3\% on average.
\\
\hspace*{5pt}
8) Results on the TASD task mirror those for E2E-ABSA: \textsc{SeqLab} consistently improves performance across settings, while \textsc{ACS} substantially enhances cross-lingual transfer. For both tasks, our best cross-lingual results are on average only 5\% weaker than monolingual performance.
\begin{table*}[ht!]
\begin{adjustbox}{width=0.95\linewidth}
\begin{tabular}{@{}llcccccccccccccccccccc@{}}
\toprule
\textbf{}               & \textbf{}                                                     & \multicolumn{10}{c}{\textbf{E2E-ABSA}}                                                                                                                                                                                                                                                & \multicolumn{10}{c}{\textbf{TASD}}                                                                                                                                                                                                                                                    \\ \cmidrule(lr){3-12} \cmidrule(lr){13-22}
\textbf{Domain}         & \textbf{Method}                                               & Es                        & Fr                        & Nl                        & Ru                        & Tr                        & Ja                        & Ko                        & Th                        & Vi                        & Zh                        & Es                        & Fr                        & Nl                        & Ru                        & Tr                        & Ja                        & Ko                        & Th                        & Vi                        & Zh                        \\ \midrule
\multirow{7}{*}{\rotatebox[origin=c]{90}{hotel}}  & \textsc{Monolingual}                                          & 67.01                     & 66.83                     & 71.89                     & 70.77                     & 69.19                     & 78.42                     & 77.85                     & 75.84                     & 70.27                     & 79.41                     & 60.02                     & 61.06                     & 63.40                     & 51.05                     & 65.10                     & 67.81                     & 68.12                     & 63.88                     & 59.64                     & 66.42                     \\
                        & \phantom{0} + \textsc{SeqLab}                                 & 69.36                     & 67.35                     & 69.98                     & 70.18                     & 70.05                     & 78.55                     & 79.26                     & 76.22                     & 71.13                     & 81.44                     & 58.12                     & 60.88                     & 65.44                     & 51.42                     & 69.51                     & 71.42                     & 71.52                     & 66.43                     & 62.49                     & 68.17                     \\ \cdashlinelr{2-22}
                        & \textsc{Zero-shot} {\footnotesize \citep{wu-etal-2025-absa}}* & 41.03 & 34.08 & 49.35 & 38.47 & 46.87 & 35.93 & 31.89 & 37.20 & 25.44 & 39.73 & 36.77 & 19.49 & 26.30 & 20.57 & 17.71 & 25.30 & 13.36 & 20.90 & 15.69 & 24.45 \\
                        & \textsc{Zero-shot}                                            & 59.66                     & 51.75                     & 57.40                     & 46.71                     & 40.21                     & 56.41                     & 31.06                     & 47.15                     & 36.28                     & 62.31                     & 55.53                     & 46.10                     & 50.59                     & 38.04                     & 38.06                     & 49.25                     & 31.23                     & 41.29                     & 30.60                     & 50.81                     \\
                        & \phantom{0} + \textsc{SeqLab}                                   & 61.52                     & 56.34                     & 57.57                     & 50.69                     & 43.60                     & 61.58                     & 33.52                     & 51.87                     & 37.57                     & 68.74                     & 56.80                     & 50.33                     & 51.17                     & 44.72                     & 38.70                     & 58.05                     & 31.44                     & 44.06                     & 35.95                     & 51.77                     \\ \cdashlinelr{2-22}
                        & \textsc{ACS}                                                  & 73.48                     & \textbf{74.29}            & \textbf{77.77}            & 73.01                     & 80.29                     & 77.46                     & 82.15                     & \textbf{77.27}            & \textbf{75.01}            & \textbf{82.97}            & 64.61                     & \textbf{69.44}            & \textbf{73.75}            & \textbf{60.59}            & \textbf{72.25}            & \textbf{73.94}            & 74.88                     & \textbf{67.81}            & 64.71                     & 69.76                     \\
                        & \phantom{0} + \textsc{SeqLab}                                          & \textbf{74.77}            & 74.04                     & 76.66                     & \textbf{74.24}            & \textbf{80.40}            & \textbf{78.05}            & \textbf{83.34}            & 76.85                     & 73.63                     & 82.82                     & \textbf{65.71}            & 66.16                     & 70.82                     & 60.15                     & 72.18                     & 70.48                     & \textbf{78.67}            & 66.29                     & \textbf{68.13}            & \textbf{70.28}            \\ \midrule
\multirow{7}{*}{\rotatebox[origin=c]{90}{laptop}} & \textsc{Monolingual}                                          & 71.68                     & 85.02                     & 75.48                     & 75.65                     & 67.54                     & 81.39                     & 79.09                     & 73.08                     & 64.91                     & 76.01                     & 58.84                     & 53.23                     & 45.15                     & 43.74                     & 48.19                     & 51.66                     & 61.01                     & 51.80                     & 49.44                     & 43.99                     \\
                        & \phantom{0} + \textsc{SeqLab}                                 & 69.96                     & 85.29                     & 74.00                     & 65.84                     & 65.84                     & 81.09                     & 79.05                     & 76.02                     & 68.24                     & 77.50                     & 59.73                     & 60.97                     & 49.28                     & 52.38                     & 48.76                     & 52.21                     & 60.04                     & 54.97                     & 51.68                     & 43.74                     \\ \cdashlinelr{2-22}
                        & \textsc{Zero-shot} {\footnotesize \citep{wu-etal-2025-absa}}* & 44.89 & 48.65 & 64.71 & 42.69 & 53.76 & 44.93 & 41.60 & 44.00 & 40.76 & 53.25 & 22.46 & 24.91 & 33.54 & 18.37 & 20.88 & 22.22 & 18.01 & 22.56 & 19.59 & 22.55 \\
                        & \textsc{Zero-shot}                                            & 40.18                     & 58.52                     & 56.18                     & 37.84                     & 51.08                     & 56.80                     & 25.95                     & 54.25                     & 40.72                     & 64.09                     & 30.87                     & 39.34                     & 33.84                     & 23.69                     & 27.30                     & 30.66                     & 17.33                     & 33.24                     & 28.55                     & 32.11                     \\
                        & \phantom{0} + \textsc{SeqLab}                                   & 44.93                     & 63.17                     & 60.85                     & 45.94                     & 54.35                     & 59.56                     & 28.62                     & 55.87                     & 43.67                     & 66.94                     & 35.41                     & 41.83                     & 36.46                     & 25.78                     & 26.90                     & 33.85                     & 24.65                     & 38.61                     & 25.64                     & 32.38                     \\ \cdashlinelr{2-22}
                        & \textsc{ACS}                                                  & 76.14                     & \textbf{87.58}            & \textbf{78.94}            & 83.84                     & 75.51                     & 81.35                     & 81.33                     & 74.21                     & 73.73                     & 77.30                     & 60.58                     & 63.41                     & 49.99                     & 56.38                     & 52.98                     & \textbf{51.72}            & \textbf{66.58}            & 55.90                     & 53.57                     & 46.07                     \\
                        & \phantom{0} + \textsc{SeqLab}                                          & \textbf{78.12}            & 86.82                     & 77.81                     & \textbf{84.56}            & \textbf{76.55}            & \textbf{81.36}            & \textbf{82.10}            & \textbf{76.51}            & \textbf{74.74}            & \textbf{79.85}            & \textbf{62.85}            & \textbf{63.47}            & \textbf{52.01}            & \textbf{58.48}            & \textbf{53.39}            & 50.98                     & 66.49                     & \textbf{57.11}            & \textbf{54.37}            & \textbf{47.61}            \\ \bottomrule 
\end{tabular}
\end{adjustbox}
\caption{Results on the E2E-ABSA and TASD tasks on laptop and hotel domains with English as the source language and other languages as the target ones. \textbf{Bold} results indicate the best outcomes for a given target language, domain, and task in cross-lingual settings. * indicates results from related work that used different task formulation.}
\label{tab:res_domains}
\end{table*}

While LLMs offer the advantage of zero- and few-shot usage without task-specific training, their performance in ABSA is often inferior to fine-tuned smaller models~\citep{zhang2023sentiment,wu2025evaluating}. Achieving competitive results typically requires fine-tuning large models, introducing substantial computational and memory overhead~\citep{icaart25}. In contrast, our framework achieves competitive and often stronger cross-lingual performance while relying on a substantially smaller model. Moreover, deploying LLMs raises practical concerns, including higher inference latency and, in the case of cloud-based models, potential data privacy issues when handling sensitive data.

\subsection{Other Domains Results}

Table~\ref{tab:res_domains} presents the results on the hotel and laptop domains with English as the source language. Similar to the restaurant domain, the results demonstrate the effectiveness of both \textsc{ACS} and \textsc{SeqLab}, with each method providing substantial improvements and their combination yielding the best cross-lingual performance. The \textsc{Zero-shot} baseline is notably weaker for Asian languages than for European ones, likely due to greater linguistic divergence from English. However, incorporating \textsc{ACS} data mitigates this gap, resulting in more consistent performance across languages.

Interestingly, the cross-lingual performance of \textsc{ACS + SeqLab} in the hotel and laptop domains often surpasses monolingual results, a trend not observed in the restaurant domain. We hypothesize that this is due to the parallel nature of these datasets: non-English training data are translations and may therefore be of lower quality. Since our approach also relies on machine translation from English, the resulting translated training data are closely aligned with the target-language training data. Furthermore, incorporating high-quality English data, along with aspect code-switched data, likely provides richer supervision, leading to improved performance compared to training solely on translated target-language data. Further analysis is provided in Appendix~\ref{app:domain}.

\subsection{Additional Results}

Appendix~\ref{app:results} provides additional results with different methods on all domains and also specifically on the restaurant domain, covering different source–target language configurations and backbone models, further validating the effectiveness of the proposed approach.

\subsection{Ablation Study}
Table~\ref{tab:ablation} presents an ablation study for the proposed approach, focusing on the E2E-ABSA task. It highlights the impact of individual components of the proposed method.

\begin{table}[ht!]
    \centering
    \begin{adjustbox}{width=0.7\linewidth}

    \begin{tabular}{@{}llccccc@{}}
    \toprule
    & \textbf{Method}        & \textbf{Es}    & \textbf{Fr}    & \textbf{Nl}    & \textbf{Ru}    & \textbf{Avg}   \\ \midrule
    \multirow{4}{*}{\rotatebox[origin=c]{90}{\small \textsc{Zero-shot}}} & \textsc{Vanilla}                & 62.30          & 57.83          & 66.46          & 60.64          & 61.81          \\
    & \textsc{SeqLab}        & 63.35          & 60.18          & 69.07          & 63.36          & 63.99          \\
    & \phantom{0}- inference & 61.19          & 58.82          & 68.37          & 63.27          & 62.91          \\
    & \textsc{Encoder-only}           & 60.89          & 55.82          & 58.47          & 52.26          & 56.86          \\ \cdashlinelr{1-7}
    \multirow{4}{*}{\rotatebox[origin=c]{90}{\textsc{ACS}}} & \textsc{Vanilla}                & 70.84          & 62.30          & 71.52          & 67.96          & 68.16          \\
    & \textsc{SeqLab}        & \textbf{72.56} & 62.42          & \textbf{72.00}          & 68.99          & \textbf{68.99}          \\
    & \phantom{0}- inference & 71.63          & \textbf{63.55} & 71.42          & 68.80          & 68.85          \\
    & \textsc{Encoder-only}           & 65.28          & 57.82          & 60.42          & 59.17          & 60.67          \\ 
    \bottomrule
    \end{tabular}
    \end{adjustbox}
    \setlength{\belowcaptionskip}{-10pt}
    \caption{Ablation study of the proposed framework on E2E-ABSA, with the best results highlighted in \textbf{bold}.}
    \label{tab:ablation}
\end{table}

\vspace{-5pt}
\paragraph{Effect of sequence labelling} The \textsc{Vanilla} approach, which excludes the sequence labelling component, performs worse in all cases. The largest difference is observed in the \textsc{Zero-shot} setting, where the \textsc{Vanilla} method lags more than 2\% behind the \textsc{SeqLab} approach.
\vspace{-5pt}
\paragraph{Effect of the inference stage} To assess the effectiveness of the inference stage, we remove it from the \textsc{SeqLab} method while retaining the sequence labelling part. In most cases, excluding the enhanced inference stage leads to worse performance, confirming its utility. However, the difference between \textsc{Vanilla} and \textsc{SeqLab} is greater than that between \textsc{SeqLab} and the \textsc{SeqLab} version without the inference stage, indicating that sequence labelling plays a more significant role than the inference stage.
\vspace{-5pt}
\paragraph{Effect of the generative framework} We evaluate only the encoder part of mT5 that uses sequence labelling with BIO tagging to predict aspect terms and polarity (\textsc{Encoder-only}). The results are 5-8\% worse than when using the full model for generating predictions, highlighting that the encoder alone underperforms compared to the full model.
\vspace{-5pt}
\paragraph{Effect of \textsc{ACS}} Incorporating \textsc{ACS} improves results by more than 5\% compared to the \textsc{Zero-shot} approach. We also evaluate a variant that excludes the aspect code-switched datasets and uses only the source and translated datasets (\textsc{Tr-only}), without the aspect-code switched data. As shown in Appendix~\ref{app:translation}, this approach yields slightly worse results on average than using \textsc{ACS}, especially in laptop and hotel domains, though it remains a viable alternative.

\begin{table}[ht!]
    \centering
    \begin{adjustbox}{width=0.5\linewidth}
    \begin{tabular}{@{}lll@{}}
    \toprule
                           & \textbf{W/o sentiment} & \textbf{W/ sentiment} \\ \midrule
    \textsc{SeqLab}                 & 80.22                  & 81.14                 \\
    \phantom{0}- inference & 79.52                  & 80.36                 \\ \bottomrule
    \end{tabular}
    \end{adjustbox}
    \setlength{\belowcaptionskip}{-10pt}
    \caption{English development set results with and without sentiment information in sequence labelling.}
    \label{tab:wo_sentiment}
\end{table}

\paragraph{Effect of sentiment information in sequence labelling} We label tokens with aspect term and sentiment polarity information. However, the sentiment polarity information is not used during the inference stage. Alternatively, we experiment with labelling tokens based solely on whether they are part of aspect terms, excluding sentiment information. Table~\ref{tab:wo_sentiment} shows the results for the English development set. Including sentiment information leads to an improvement of almost 1\%, demonstrating its effectiveness as it contributes to the model's overall understanding of the sentiment elements.

\subsubsection{Loss and Threshold Hyperparameters}
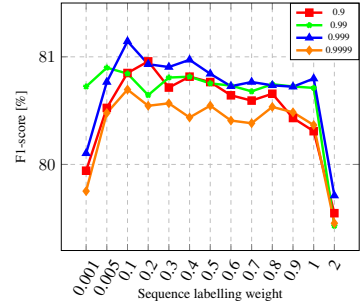
\begin{figure}[ht!]
    \centering
        \centering
        \begin{adjustbox}{width=0.6\linewidth}
            \begin{tikzpicture}
                \begin{axis}[
                    xlabel={\small{Sequence labelling weight}},
                    ylabel={\small{F1-score [\%]}},
                    xlabel near ticks,
                    xlabel style={at={(axis description cs:0.5,-0.13)}, anchor=north},
                    symbolic x coords={0.001, 0.005, 0.1, 0.2, 0.3, 0.4, 0.5, 0.6, 0.7, 0.8, 0.9, 1, 2},
                    xtick=data,
                    ymin=79.2, ymax=81.5,
                    ytick={80,81},
                    legend pos=north east,
                    ymajorgrids=true,
                    xmajorgrids=true,
                    grid style=dashed,
                    legend style={at={(1,1)}, anchor=north east, font=\footnotesize, nodes={scale=0.7, transform shape}},
                    xticklabel style={rotate=60},
                    ]

                    \addplot[
                        color=red,
                        mark=square*,
                        ultra thick,
                    ] 
                    coordinates {  
                        (0.001,79.94160818)  
                        (0.005,80.5251097)  
                        (0.1,80.84957298)  
                        (0.2,80.95763724)  
                        (0.3,80.71429512)  
                        (0.4,80.81560467)  
                        (0.5,80.76542554)  
                        (0.6,80.64219562)  
                        (0.7,80.5932486)  
                        (0.8,80.65681398)  
                        (0.9,80.43132817)  
                        (1,80.30884444)  
                        (2,79.54691515)  
                    };  
                    
                    \addlegendentry{0.9}

                    \addplot[
                        color=green,
                        mark=star,
                        ultra thick,
                    ]
                    coordinates {  
                        (0.001,80.72355879)  
                        (0.005,80.89890931)  
                        (0.1,80.84395576)  
                        (0.2,80.64402109)  
                        (0.3,80.80722119)  
                        (0.4,80.81553686)  
                        (0.5,80.7530959)  
                        (0.6,80.73281787)  
                        (0.7,80.6792338)  
                        (0.8,80.74505016)  
                        (0.9,80.7243597)  
                        (1,80.71111762)  
                        (2,79.42951776)  
                    };

                    \addlegendentry{0.99}
    
                    \addplot[
                        color=blue,
                        mark=triangle*,
                        ultra thick,
                    ] 
                    coordinates {
                        (0.001,80.1035533)
                        (0.005,80.76571214)
                        (0.1,81.14407063)
                        (0.2,80.93119041)
                        (0.3,80.90497851)
                        (0.4,80.97226472)
                        (0.5,80.84234731)
                        (0.6,80.72684665)
                        (0.7,80.76575792)
                        (0.8,80.73521731)
                        (0.9,80.7222877)
                        (1,80.7978183)
                        (2,79.71056747)
                    };

                    \addlegendentry{0.999}

                    \addplot[
                        color=orange,
                        mark=diamond*,
                        ultra thick,
                    ] 
                    coordinates {  
                        (0.001,79.75187833)  
                        (0.005,80.47748748)  
                        (0.1,80.69561925)  
                        (0.2,80.54500551)  
                        (0.3,80.56786664)  
                        (0.4,80.4363019)  
                        (0.5,80.54614779)  
                        (0.6,80.40766859)  
                        (0.7,80.38267631)  
                        (0.8,80.53461321)  
                        (0.9,80.48435491)  
                        (1,80.36578876)  
                        (2,79.4538171)  
                    };  

                    \addlegendentry{0.9999}
                \end{axis}
            \end{tikzpicture}
	   \end{adjustbox}
       \caption{Results with different sequence labelling weights and thresholds on the English development set.}
    \label{fig:hyperparameters}
\end{figure}

Figure~\ref{fig:hyperparameters} investigates the impact of the sequence labelling weight ($\alpha$) and threshold parameters on the English development set. We evaluate a range of sequence labelling weights. The best results are generally achieved with $\alpha = 0.1$, but any value between 0.005 and 1 performs similarly well. Setting $\alpha$ too high (e.g. 2) significantly reduces performance, indicating that the generation loss remains critical. Conversely, a very low value (0.001) also lowers performance, suggesting that the sequence labelling task plays an important role.

For the threshold value, we experiment with four different settings. Performance remains relatively stable across thresholds, but as the threshold increases, it generally improves up to 0.999. However, using 0.9999 results in the worst performance. The optimal threshold of 0.999 is close to 1, and when examining the probabilities of generated words, we find they often approach 1, even for incorrect predictions. Therefore, 0.999 is a reasonable choice, as smaller thresholds may allow incorrect outputs to pass through, while larger thresholds could be too restrictive.

\begin{table}[ht!]
\centering
\begin{adjustbox}{width=0.9\linewidth}
\begin{tabular}{@{}rrrrrrrrr@{}}
\toprule
\textbf{Threshold} & \textbf{F1} & \textbf{\begin{tabular}[c]{@{}r@{}}Total \\ predictions\end{tabular}} & \textbf{\begin{tabular}[c]{@{}r@{}}Accepted \\ (threshold)\end{tabular}} & \textbf{\begin{tabular}[c]{@{}r@{}}Accepted\\ (encoder)\end{tabular}} & \textbf{\begin{tabular}[c]{@{}r@{}}Ratio accepted\\ threshold [\%]\end{tabular}} & \textbf{\begin{tabular}[c]{@{}r@{}}Ratio acepted\\ encoder [\%]\end{tabular}} & \textbf{Unaccepted} & \textbf{\begin{tabular}[c]{@{}r@{}}Unaccepted\\  ratio [\%]\end{tabular}} \\ \midrule
0.9                & 69.77       & 373                                                                   & 268                                                                      & 80                                                                    & 71.85                                                                            & 21.45                                                                         & 25                  & 6.70                                                                      \\
0.99               & 69.73       & 326                                                                   & 198                                                                      & 101                                                                   & 60.74                                                                            & 30.98                                                                         & 27                  & 8.28                                                                      \\
0.999              & 72.45       & 387                                                                   & 39                                                                       & 291                                                                   & 10.08                                                                            & 75.19                                                                        & 57                  & 14.73                                                                     \\
0.9999             & 71.21       & 410                                                                   & 2                                                                        & 333                                                                   & 0.49                                                                             & 81.22                                                                         & 75                  & 18.29                                                                     \\ \bottomrule
\end{tabular}
\end{adjustbox}
\caption{Effect of threshold on accepted predictions for Spanish in the E2E-ABSA task and restaurant dataset.}

\label{tab:threshold-analysis}
\end{table}

\begin{figure*}[ht!]
    \begin{subfigure}{0.32\textwidth}
        \centering
        \begin{adjustbox}{width=\linewidth}
            \begin{tikzpicture}
                \begin{axis}[
                    ybar,
                    bar width=6pt,
                    xtick={0,1,2,3, 4},
                    ytick={0,10,20,30,40,50,60},
                    xticklabels={Es, Fr, Nl, Ru, Tr},
                    ymin=0,
                    ymax=70,
                    xmin=-0.5,
                    xmax=4.5,
                    ymajorgrids=true,
                    ylabel={\footnotesize Number of errors},
                    xlabel={\footnotesize Target language},
                    legend style={at={(0.39,1)}, anchor=north west, font=\footnotesize, nodes={scale=0.7, transform shape}},
                    ]
                    \addplot[black,fill=lightblue,postaction={pattern=north west lines}] coordinates {(0,49) (1,64) (2,43) (3,45) (4,64)};
                    \addplot[black,fill=lightorange,postaction={pattern=north east lines}] coordinates {(0,41) (1,61) (2,42) (3,36) (4,59)};
                    \addplot[black,fill=lightgreen,postaction={pattern=grid}] coordinates {(0,37) (1,55) (2,38) (3,33) (4,58)};
                    \addplot[black,fill=lightpurple,postaction={pattern=crosshatch dots}] coordinates {(0,33) (1,49) (2,37) (3,29) (4,56)};
                    \legend{\textsc{Zero-shot}, \textsc{Zero-shot + SeqLab}, \textsc{ACS}, \textsc{ACS + SeqLab}}
                \end{axis}
            \end{tikzpicture}
        \end{adjustbox}
       \subcaption{Aspect term}
    \end{subfigure}
    \hfill
    \begin{subfigure}{0.32\textwidth}
        \centering
        \begin{adjustbox}{width=\linewidth}
            \begin{tikzpicture}
                \begin{axis}[
                    ybar,
                    bar width=6pt,
                    xtick={0,1,2,3, 4},
                    ytick={0,10,20,30,40},
                    xticklabels={Es, Fr, Nl, Ru, Tr},
                    ymin=0,
                    ymax=50,
                    xmin=-0.5,
                    xmax=4.5,
                    ymajorgrids=true,
                    ylabel={\footnotesize Number of errors},
                    xlabel={\footnotesize Target language},
                    legend style={at={(0.39,1)}, anchor=north west, font=\footnotesize, nodes={scale=0.7, transform shape}},
                    ]
                    \addplot[black,fill=lightblue,postaction={pattern=north west lines}] coordinates {(0,31) (1,40) (2,31) (3,23) (4,45)};
                    \addplot[black,fill=lightorange,postaction={pattern=north east lines}] coordinates {(0,23) (1,41) (2,30) (3,20) (4,43)};
                    \addplot[black,fill=lightgreen,postaction={pattern=grid}] coordinates {(0,18) (1,37) (2,28) (3,17) (4,36)};
                    \addplot[black,fill=lightpurple,postaction={pattern=crosshatch dots}] coordinates {(0,14) (1,36) (2,26) (3,15) (4,38)};
                    \legend{\textsc{Zero-shot}, \textsc{Zero-shot + SeqLab}, \textsc{ACS}, \textsc{ACS + SeqLab}}
                \end{axis}
            \end{tikzpicture}
        \end{adjustbox}
       \subcaption{Aspect category}
    \end{subfigure}
    \hfill
    \begin{subfigure}{0.32\textwidth}
        \centering
        \begin{adjustbox}{width=\linewidth}
            \begin{tikzpicture}
                \begin{axis}[
                    ybar,
                    bar width=6pt,
                    xtick={0,1,2,3, 4},
                    ytick={0,10,20,30},
                    xticklabels={Es, Fr, Nl, Ru, Tr},
                    ymin=0,
                    ymax=40,
                    xmin=-0.5,
                    xmax=4.5,
                    ymajorgrids=true,
                    ylabel={\footnotesize Number of errors},
                    xlabel={\footnotesize Target language},
                    legend style={at={(0.39,1)}, anchor=north west, font=\footnotesize, nodes={scale=0.7, transform shape}},
                    ]
                    \addplot[black,fill=lightblue,postaction={pattern=north west lines}] coordinates {(0,16) (1,24) (2,15) (3,15) (4,36)};
                    \addplot[black,fill=lightorange,postaction={pattern=north east lines}] coordinates {(0,11) (1,23) (2,14) (3,13) (4,32)};
                    \addplot[black,fill=lightgreen,postaction={pattern=grid}] coordinates {(0,10) (1,23) (2,14) (3,14) (4,28)};
                    \addplot[black,fill=lightpurple,postaction={pattern=crosshatch dots}] coordinates {(0,10) (1,22) (2,13) (3,13) (4,24)};
                    \legend{\textsc{Zero-shot}, \textsc{Zero-shot + SeqLab}, \textsc{ACS}, \textsc{ACS + SeqLab}}
                \end{axis}
            \end{tikzpicture}
        \end{adjustbox}
        \subcaption{Sentiment polarity}
    \end{subfigure}
    \setlength{\belowcaptionskip}{-6pt}
    \caption{Number of error types for the TASD task in restaurant domain for each target language with English as the source language.}
    \label{fig:error}
\end{figure*}
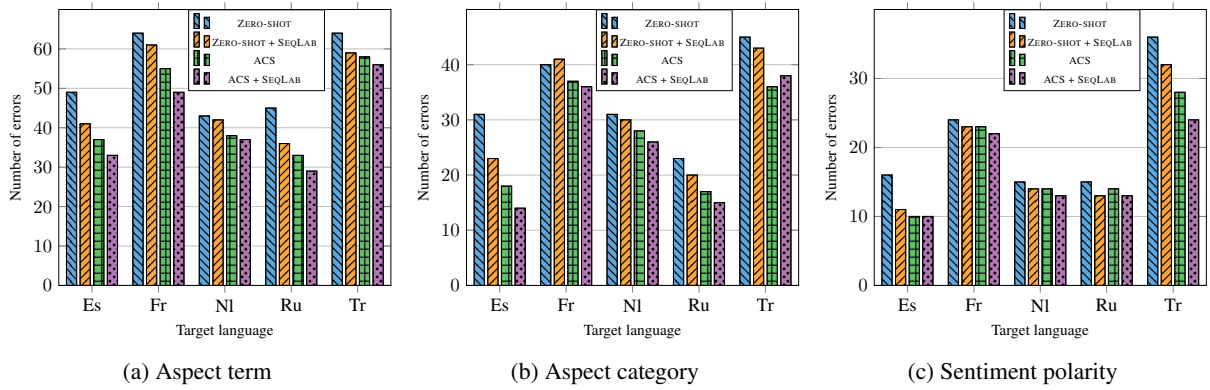

To further justify our choice of threshold, we analyse how many predictions are accepted by the thresholding mechanism versus by the encoder filtering. Table~\ref{tab:threshold-analysis} shows the breakdown for Spanish on the E2E-ABSA task (other languages and TASD show the same pattern, with similar shifts in the ratios of threshold-accepted, encoder-accepted and unaccepted predictions as the threshold increases). Lower thresholds (0.9--0.99) admit a high proportion of low-confidence predictions (over 60--70\%), which correlates with lower F1 scores. As the threshold increases, the number of threshold-accepted predictions decreases sharply, shifting acceptance to the encoder filter. The threshold of 0.999 achieves the best balance: it suppresses most low-confidence outputs while still allowing enough high-quality predictions to pass, yielding the highest F1 (72.45). In contrast, 0.9999 becomes overly restrictive, accepting almost no predictions directly and resulting in a performance drop. Overall, the table illustrates that 0.999 provides the most effective trade-off between precision and coverage.

\subsubsection{Error Analysis}

We conduct an error analysis to assess the performance of different methods across target languages and identify key challenges. We randomly select 100 test examples for each target language and evaluate all methods using the same examples. We use English as the source language for the TASD task and manually compare model outputs to the correct labels. Figure~\ref{fig:error} presents the results, while Appendix~\ref{app:qualitative} shows a small set of illustrative model errors. As expected, the \textsc{Zero-shot} approach produces the most errors, followed by \textsc{Zero-shot + SeqLab} and \textsc{ACS}, while \textsc{ACS + SeqLab} consistently has the fewest errors.

Predicting aspect terms is the most challenging, as they are language-dependent and can be any word or phrase in the text. Errors include missing aspect terms, incorrect spans, and partial matches (missing or adding words). The \textsc{Zero-shot} method occasionally predicts aspect terms in English rather than the target language (e.g. \textit{\quotes{staff}} instead of French \textit{\quotes{personnel}}), particularly in Spanish, French, and Turkish. \textsc{SeqLab} and \textsc{ACS} mitigate these issues, as they are primarily designed to improve aspect term prediction. Missing aspect terms are especially common in French. 

Aspect category prediction is comparatively easier than aspect term prediction due to the limited label space. The main challenges arise with semantically similar categories, such as \textit{\quotes{food quality}} and \textit{\quotes{food style\_options}}, and rare categories like \textit{\quotes{location general}}. Although \textsc{SeqLab} does not explicitly model aspect categories, it still reduces errors, suggesting that better aspect term understanding and prediction enhances performance even for other sentiment elements.

Sentiment polarity is the easiest element to predict, with most errors involving the \textit{\quotes{neutral}} class. The model often misclassifies mildly positive or negative sentiment as \textit{\quotes{positive}} or \textit{\quotes{negative}} instead of \textit{\quotes{neutral}}. Additionally, \textit{\quotes{neutral}} is the least frequent sentiment class in all datasets, making it harder to learn. While \textsc{SeqLab} and \textsc{ACS} reduce polarity errors, the effect is smaller than for aspect terms or categories.

\subsection{Machine Translation Quality Analysis}
\label{app:translation}
To assess translation quality, we evaluate translations across all languages and domains using all training examples translated from English, excluding cases where the translation omitted the special markers around aspect terms, which occurred in approximately 6\% of instances. We report three metrics commonly used in prior ABSA studies to evaluate translation quality~\citep{wu-etal-2025-absa}.

\begin{table*}[ht!]
\centering
\begin{adjustbox}{width=\linewidth}
\begin{tabular}{@{}lccccccccccccccccccccc@{}}
\toprule
   & \multicolumn{7}{c}{\textbf{Restaurant}}                                                                                         & \multicolumn{7}{c}{\textbf{Hotel}}                                                                                              & \multicolumn{7}{c}{\textbf{Laptop}}                                                                                             \\ \cmidrule(lr){2-8} \cmidrule(lr){9-15} \cmidrule(lr){16-22}
   & \multicolumn{3}{c}{\textbf{$\text{Tr}\leftrightarrow\text{Tr}_\text{w/o markers}$}} & \multicolumn{2}{c}{\textbf{$\text{En}\leftrightarrow\text{Tr}$}} & \multicolumn{2}{c}{\textbf{$\text{En}\leftrightarrow\text{BTr}$}}& \multicolumn{3}{c}{\textbf{$\text{Tr}\leftrightarrow\text{Tr}_\text{w/o markers}$}} & \multicolumn{2}{c}{\textbf{$\text{En}\leftrightarrow\text{Tr}$}} & \multicolumn{2}{c}{\textbf{$\text{En}\leftrightarrow\text{BTr}$}}& \multicolumn{3}{c}{\textbf{$\text{Tr}\leftrightarrow\text{Tr}_\text{w/o markers}$}} & \multicolumn{2}{c}{\textbf{$\text{En}\leftrightarrow\text{Tr}$}} & \multicolumn{2}{c}{\textbf{$\text{En}\leftrightarrow\text{BTr}$}} \\ \cmidrule(lr){2-4} \cmidrule(lr){5-6} \cmidrule(lr){7-8} \cmidrule(lr){9-11} \cmidrule(lr){12-13} \cmidrule(lr){14-15} \cmidrule(lr){16-18} \cmidrule(lr){19-20} \cmidrule(lr){21-22}    & \textbf{Acc}          & \textbf{BS}          & \textbf{SB}          & \textbf{BS}                  & \textbf{SB}                 & \textbf{BS}                  & \textbf{SB}                  & \textbf{Acc}          & \textbf{BS}          & \textbf{SB}          & \textbf{BS}                  & \textbf{SB}                 & \textbf{BS}                  & \textbf{SB}                  & \textbf{Acc}          & \textbf{BS}          & \textbf{SB}          & \textbf{BS}                  & \textbf{SB}                 & \textbf{BS}                  & \textbf{SB}                  \\ \midrule
Es & 86.62        & 98.73       & 99.41       & 92.20               & 93.14              & 96.97               & 97.33               & 93.64        & 98.80       & 99.53       & 92.62               & 93.46              & 96.99               & 97.61               & 83.01        & 98.48       & 99.19       & 92.29               & 93.53              & 96.43               & 97.37               \\
Fr & 80.70        & 98.45       & 99.39       & 91.99               & 92.12              & 97.05               & 97.42               & 85.02        & 98.70       & 99.44       & 92.01               & 92.70              & 96.99               & 97.59               & 80.60        & 98.62       & 99.46       & 91.77               & 93.02              & 96.56               & 97.39               \\
Nl & 75.32        & 98.59       & 99.35       & 92.77               & 92.66              & 97.38               & 97.61               & 91.89        & 99.07       & 99.54       & 93.11               & 93.03              & 97.38               & 97.85               & 72.89        & 98.32       & 99.05       & 92.95               & 93.31              & 96.95               & 97.57               \\
Ru & 76.25        & 98.43       & 99.16       & 91.53               & 92.10              & 96.55               & 97.02               & 73.06        & 98.29       & 99.17       & 91.83               & 92.78              & 96.57               & 97.23               & 75.13        & 98.13       & 98.97       & 92.03               & 93.29              & 95.96               & 96.89               \\
Tr & 68.65        & 97.72       & 98.87       & 90.35               & 91.46              & 96.64               & 97.08               & 86.42        & 98.25       & 99.24       & 90.75               & 92.36              & 96.65               & 97.33               & 75.17        & 97.84       & 98.89       & 90.68               & 92.43              & 95.98               & 97.05               \\
Ja  & --           & --          & --          & --                  & --                 & --                  & --                  & 73.90        & 97.15       & 98.51       & 90.11               & 91.16              & 95.93               & 96.87               & 76.18        & 97.01       & 98.60       & 90.05               & 91.95              & 95.22               & 96.59               \\
Ko  & --           & --          & --          & --                  & --                 & --                  & --                  & 72.74        & 97.51       & 98.83       & 90.29               & 90.58              & 96.10               & 96.94               & 76.21        & 97.45       & 98.69       & 90.26               & 91.44              & 95.49               & 96.71               \\
Vi  & --           & --          & --          & --                  & --                 & --                  & --                  & 83.71        & 98.12       & 98.92       & 91.09               & 91.43              & 96.50               & 97.04               & 80.00        & 97.99       & 98.69       & 91.47               & 91.75              & 96.02               & 96.96               \\
Th  & --           & --          & --          & --                  & --                 & --                  & --                  & 78.13        & 95.80       & 98.01       & 90.07               & 91.77              & 95.74               & 96.52               & 77.47        & 95.60       & 97.81       & 90.29               & 92.64              & 95.32               & 96.59               \\
Zh  & --           & --          & --          & --                  & --                 & --                  & --                  & 72.25        & 98.22       & 99.31       & 91.22               & 89.74              & 96.03               & 96.47               & 77.44        & 97.07       & 98.35       & 91.10               & 91.33              & 95.77               & 96.97               \\ \cdashlinelr{1-22}
AVG & 77.51        & 98.38       & 99.24       & 91.77               & 92.30              & 96.92               & 97.29               & 81.08        & 97.99       & 99.05       & 91.31               & 91.90              & 96.49               & 97.15               & 77.41        & 97.65       & 98.77       & 91.29               & 92.47              & 95.97               & 97.01              
\\ \bottomrule        
\end{tabular}
\end{adjustbox}
\caption{Evaluation of translation quality from English to target languages using multiple metrics. \textit{Acc} denotes Aspect accuracy, \textit{BS} BERTScore, and \textit{SB} SBERT similarity. \textit{Tr} refers to translated text, \textit{Tr}\textsubscript{w/o markers} to translations generated without special markers around aspect terms, \textit{En} to the original English sentence, and \textit{BTr} to the back-translation into English.}
\label{tab:tr}
\end{table*}

The first metric is \textit{Aspect accuracy} (\textit{Acc}), which measures the percentage of aspect terms correctly preserved in the translation\textsubscript{w/o markers}, i.e. translations generated without special markers around aspect terms. The second metric is BERTScore~\citep{2020BERTScore} computed with XLM-R-large, and the third is the SBERT score~\citep{reimers-gurevych-2019-sentence} computed using multilingual E5~\citep{wang2024multilinguale5textembeddings}, where sentence embeddings are compared using cosine similarity. BERTScore and SBERT capture semantic similarity at the token and sentence levels, respectively. For BERTScore and SBERT, we evaluate three comparison settings: (1) translated text vs. translation\textsubscript{w/o markers}, (2) translated text vs. the original English sentence in the multilingual setup, and (3) the original English sentence vs. the back-translation of the translated sentence into English. These comparisons assess translation faithfulness, i.e. whether the translated sentences preserve the meaning of the original English input. Scores above 90\% generally indicate very high semantic similarity.

Table~\ref{tab:tr} presents the translation quality analysis. Both BERTScore and SBERT remain consistently high across all languages and domains, indicating strong semantic consistency between the original English sentences and their translations. On average, the comparison between translated text and translation\textsubscript{w/o markers} achieves over 97\% BERTScore and 98\% SBERT similarity. In the multilingual comparison between translated and original English sentences, both metrics exceed 91\% on average. Slightly lower scores are expected in this setting, as the comparison is performed across different languages. For the comparison between original and back-translated sentences, both metrics exceed 95\% on average, further confirming strong semantic preservation. Overall, Asian languages tend to score approximately 0.5--1\% lower than European languages.

Aspect accuracy is also consistently high across languages, averaging approximately 77--80\% across domains. This metric is based on strict exact string matching and therefore penalizes benign linguistic variations, such as valid synonym substitutions or minor word-order changes. A manual analysis of mismatched cases confirmed that these variations generally preserve the original meaning and sentiment, consistent with the high BERTScore and SBERT results. Thus, the lower exact-match accuracy does not necessarily translate into substantial semantic noise in the pseudo-labels. Similar to the semantic similarity metrics, \textit{Acc} is generally slightly lower for Asian languages than for European languages, which is expected given the greater linguistic distance from English.

\section{Conclusion}

In this paper, we propose a novel approach to improving cross-lingual ABSA by leveraging a sequence-to-sequence model augmented with an auxiliary sequence-labelling task. This auxiliary task enhances aspect term recognition and refines sentiment predictions. Additionally, we combine this approach with aspect-code switching, a translation-based technique that swaps aspect terms between source and target sentences, generating additional training examples to improve cross-lingual generalization. Our method achieves new state-of-the-art results on the E2E-ABSA and TASD tasks, demonstrating its effectiveness across eleven languages, three domains, and two backbone models. Furthermore, we systematically evaluate different source-target language pairs. Finally, our error analysis highlights key challenges, offering insights for future research.

\section*{Limitations}

Despite achieving state-of-the-art performance in cross-lingual ABSA, the proposed framework has some limitations. First, while the experiments confirmed its effectiveness for cross-lingual ABSA, it could be extended to tasks like named entity recognition. Second, our method adds a simple linear layer on top of the encoder that is not present in traditional sequence-to-sequence models, though this overhead is very small. Third, we use a relatively simple sequence labelling approach that does not encode information about aspect categories and operates at the level of individual words rather than word spans, while aspect terms are often multi-word. A more expressive labelling scheme could potentially improve performance. Fourth, we focus exclusively on encoder–decoder models and do not explore decoder-only alternatives, which have recently gained popularity. However, their unidirectional nature and lack of fine-grained token-level supervision may make them less suitable for sequence labelling tasks.

\section*{Ethics Statement}
We base our experiments on widely adopted benchmark datasets from prior research, allowing for fair and transparent result comparison. The study was conducted ethically and does not involve any harm to individuals. Nonetheless, the pre-trained models used may reflect unintended biases -- such as those related to race or gender -- due to the nature of the large-scale web data on which they were trained.

\section*{Acknowledgements}
This work has been supported by the Grant No. SGS-2025-022 -- New Data Processing Methods in Current Areas of Computer Science.
Computational resources were provided by the e-INFRA CZ project (ID:90254), supported by the Ministry of Education, Youth and Sports of the Czech Republic.

\bibliography{bibliography}

\appendix

\section{Dataset Statistics}
\label{app:data_stats}
 Table~\ref{tab:data} shows the statistics of the restaurant dataset for each language. Table~\ref{tab:data_domains} presents the statistics for the hotel and laptop domains, where the number of sentences and aspects is identical across languages due to the parallel nature of the dataset.
\begin{table}[ht!]
    \centering
    \begin{adjustbox}{width=0.95\linewidth}   
        \begin{tabular}{@{}llrrrrrr@{}}
        \toprule
                               &     & \textbf{En} & \textbf{Es} & \textbf{Fr} & \textbf{Nl} & \textbf{Ru} & \textbf{Tr} \\ \midrule
        \multirow{2}{*}{Train} & No. sentences & 1,600       & 1,656       & 1,332       & 1,378       & 2,924 & 986      \\
                               & No. aspects & 1,377       & 1,500       & 1,294       & 956         & 2,439 & 1,083       \\ \cdashlinelr{1-8}
        \multirow{2}{*}{Dev}   & No. sentences & 400         & 414         & 322        & 344         & 731 & 246        \\
                               & No. aspects & 365         & 353         & 345         & 274         & 629   & 271      \\\cdashlinelr{1-8}
        \multirow{2}{*}{Test}  & No. sentences & 676         & 881         & 668         & 575         & 1,209   & 144    \\
                               & No. aspects & 612         & 713         & 649         & 373         & 945   & 148      \\ \bottomrule 
        \end{tabular}
    \end{adjustbox}
    \caption{Data statistics for each language in the restaurant domain.}
    \label{tab:data}
\end{table}

\begin{table}[ht!]
    \centering
    \begin{adjustbox}{width=0.6\linewidth}   
        \begin{tabular}{@{}llrr@{}}
        \toprule
                               &     & \textbf{Hotel} & \textbf{Laptop} \\ \midrule
        \multirow{2}{*}{Train} & No. sentences & 1,255       & 1,264        \\
                               & No. aspects & 1,250       & 1,366           \\ \cdashlinelr{1-4}
        \multirow{2}{*}{Dev}   & No. sentences & 308         & 326               \\
                               & No. aspects & 280         & 352            \\\cdashlinelr{1-4}
        \multirow{2}{*}{Test}  & No. sentences & 584         & 532            \\
                               & No. aspects & 714         & 610           \\ \bottomrule 
        \end{tabular}
    \end{adjustbox}
    \caption{Data statistics for hotel and laptop domains.}
    \label{tab:data_domains}
\end{table}

\section{Experimental Details}
\label{app:details}
For all experiments, we use models from the HuggingFace Transformers library\footnote{\url{https://github.com/huggingface/transformers}}~\citep{wolf-etal-2020-transformers}, the AdamW optimizer~\citep{loshchilov2019decoupledweightdecayregularization}, a batch size of 16, greedy search decoding, an inference threshold of 0.999, a sequence labelling loss weight $\alpha$ of 0.1, and a single NVIDIA L40 GPU with 48 GB memory. We search for the optimal number of epochs up to a maximum of 25.

We employ the base mT5 model~\citep{xue-etal-2021-mt5} for the main experiments with a learning rate of 1e-4, as monolingual studies have successfully used English base T5~\citep{t5}. To assess the robustness of our method, we also conduct experiments with the large mBART model~\cite{tang2020multilingual} (learning rate 1e-5) and the large mT5 model (learning rate 1e-4).

We use the Google Translate API\footnote{\url{https://translate.google.com/}} for the translation process described in Section~\ref{sec:acs} but discard approximately 6\% of cases where special symbols are lost to maintain data integrity. Figure~\ref{fig:acs} illustrates the process. \citet{zhang-etal-2021-cross} provide \textsc{ACS} datasets with English as the source language and various target languages, excluding Turkish. However, they removed aspect category annotations, preventing their use for TASD. To address this, we recover the original aspect categories and generate new datasets for Turkish as a target language and for all source languages other than English.

\begin{figure}[ht!]
    \centering
    \includegraphics[width=0.93\linewidth]{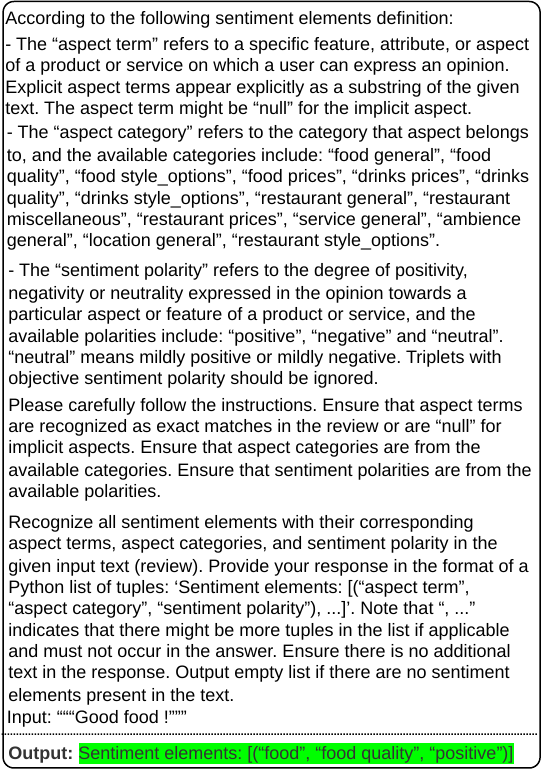}
    \caption{Prompt for the TASD task with task definition, example input and expected output (green box).}
    \label{fig:prompt}
\end{figure}

\begin{table*}[ht!]
\centering
\begin{adjustbox}{width=\linewidth}
\begin{tabular}{@{}llcccccccccccccccccccc@{}}
\toprule
                             &                               & \multicolumn{10}{c}{\textbf{E2E-ABSA}}                                                                                                                                        & \multicolumn{10}{c}{\textbf{TASD}}                                                                                                                                            \\ \cmidrule(lr){3-12} \cmidrule(lr){13-22}
\textbf{Domain}              & \textbf{Method}               & Es             & Fr             & Nl             & Ru             & Tr                   & Ja             & Ko             & Th             & Vi             & Zh             & Es             & Fr             & Nl             & Ru             & Tr                   & Ja             & Ko             & Th             & Vi             & Zh             \\ \midrule
\multirow{10}{*}{\rotatebox[origin=c]{90}{hotel}}      & \textsc{Monolingual}          & 67.01          & 66.83          & 71.89          & 70.77          & 69.19                & 78.42          & 77.85          & 75.84          & 70.27          & 79.41          & 60.02          & 61.06          & 63.40          & 51.05          & 65.10                & 67.81          & 68.12          & 63.88          & 59.64          & 66.42          \\
                             & \phantom{0} + \textsc{SeqLab} & 69.36          & 67.35          & 69.98          & 70.18          & 70.05                & 78.55          & 79.26          & 76.22          & 71.13          & 81.44          & 58.12          & 60.88          & 65.44          & 51.42          & 69.51                & 71.42          & 71.52          & 66.43          & 62.49          & 68.17          \\ \cdashlinelr{2-22}
                             & \textsc{Multilingual}         & 72.06          & 72.23          & 74.17          & 69.34          & 77.27                & 77.95          & 77.59          & 73.85          & 70.90          & 79.89          & 63.39          & 68.22          & 68.73          & 59.60          & 69.36                & \textbf{75.18} & 73.61          & 64.98          & 60.78          & 68.75          \\
                             & \phantom{0} + \textsc{SeqLab} & 73.50          & 70.68          & 73.16          & 71.88          & 77.93                & 79.73          & 80.15          & 76.67          & 72.36          & 82.94          & 62.45          & 68.27          & 69.44          & \textbf{61.17} & 70.92                & 72.31          & 75.57          & 66.21          & 65.03          & \textbf{70.85} \\ \cdashlinelr{2-22}
                             & \textsc{Zero-shot}             & 59.66          & 51.75          & 57.40          & 46.71          & 40.21                & 56.41          & 31.06          & 47.15          & 36.28          & 62.31          & 55.53          & 46.10          & 50.59          & 38.04          & 38.06                & 49.25          & 31.23          & 41.29          & 30.60          & 50.81          \\
                             & \phantom{0} + \textsc{SeqLab} & 61.52          & 56.34          & 57.57          & 50.69          & 43.60                & 61.58          & 33.52          & 51.87          & 37.57          & 68.74          & 56.80          & 50.33          & 51.17          & 44.72          & 38.70                & 58.05          & 31.44          & 44.06          & 35.95          & 51.77          \\ \cdashlinelr{2-22}
                             & \textsc{Tr-only}              & 73.07          & 72.99          & 76.48          & 69.87          & 78.59                & \textbf{80.34} & 78.55          & 76.36          & 72.65          & 80.89          & 64.47          & 67.57          & 69.92          & 59.66          & 69.04                & 73.14          & 71.30          & 65.21          & 61.16          & 67.86          \\
                             & \phantom{0} + \textsc{SeqLab} & \textbf{74.77} & 71.19          & 73.15          & 72.25          & 77.33                & 77.78          & 81.20          & 76.87          & 72.42          & 82.36          & \textbf{66.04} & 69.34          & 68.29          & 59.91          & 68.83                & 71.72          & 74.54          & 66.03          & 63.74          & 70.62          \\ \cdashlinelr{2-22}
                             & \textsc{ACS}                  & 73.48          & \textbf{74.29} & \textbf{77.77} & 73.01          & 80.29                & 77.46          & 82.15          & \textbf{77.27} & \textbf{75.01} & \textbf{82.97} & 64.61          & \textbf{69.44} & \textbf{73.75} & 60.59          & \textbf{72.25}       & 73.94          & 74.88          & \textbf{67.81} & 64.71          & 69.76          \\
                             & \phantom{0} + \textsc{SeqLab} & \textbf{74.77} & 74.04          & 76.66          & \textbf{74.24} & \textbf{80.40}       & 78.05          & \textbf{83.34} & 76.85          & 73.63          & 82.82          & 65.71          & 66.16          & 70.82          & 60.15          & 72.18                & 70.48          & \textbf{78.67} & 66.29          & \textbf{68.13} & 70.28          \\ \midrule
\multirow{10}{*}{\rotatebox[origin=c]{90}{laptop}}     & \textsc{Monolingual}          & 71.68          & 85.02          & 75.48          & 75.65          & 67.54                & 81.39          & 79.09          & 73.08          & 64.91          & 76.01          & 58.84          & 53.23          & 45.15          & 43.74          & 48.19                & 51.66          & 61.01          & 51.80          & 49.44          & 43.99          \\
                             & \phantom{0} + \textsc{SeqLab} & 69.96          & 85.29          & 74.00          & 65.84          & 65.84                & 81.09          & 79.05          & 76.02          & 68.24          & 77.50          & 59.73          & 60.97          & 49.28          & 52.38          & 48.76                & \textbf{52.21} & 60.04          & 54.97          & 51.68          & 43.74          \\ \cdashlinelr{2-22}
                             & \textsc{Multilingual}         & 72.36          & 83.71          & 72.25          & 77.74          & 68.99                & 82.46          & 78.47          & 71.84          & 66.65          & 75.40          & 57.82          & 58.25          & 46.87          & 50.07          & 51.86                & 50.70          & 62.13          & 53.65          & 51.08          & 45.31          \\
                             & \phantom{0} + \textsc{SeqLab} & 74.11          & 85.84          & 74.85          & 81.79          & 72.96                & \textbf{83.31} & 78.84          & 76.36          & 73.94          & 79.49          & 58.42          & 62.39          & 48.75          & 56.57          & 51.43                & 51.90          & 63.30          & 53.81          & 52.64          & 46.02          \\ \cdashlinelr{2-22}
                             & \textsc{Zero-shot}             & 40.18          & 58.52          & 56.18          & 37.84          & 51.08                & 56.80          & 25.95          & 54.25          & 40.72          & 64.09          & 30.87          & 39.34          & 33.84          & 23.69          & 27.30                & 30.66          & 17.33          & 33.24          & 28.55          & 32.11          \\
                             & \phantom{0} + \textsc{SeqLab} & 44.93          & 63.17          & 60.85          & 45.94          & 54.35                & 59.56          & 28.62          & 55.87          & 43.67          & 66.94          & 35.41          & 41.83          & 36.46          & 25.78          & 26.90                & 33.85          & 24.65          & 38.61          & 25.64          & 32.38          \\ \cdashlinelr{2-22}
                             & \textsc{Tr-only}              & 72.64          & 83.86          & 72.93          & 77.90          & 72.04                & 81.00          & 78.16          & 72.73          & 69.73          & 73.95          & 57.53          & 59.15          & 47.12          & 50.64          & 48.08                & 50.41          & 60.15          & 53.26          & 51.70          & 45.82          \\
                             & \phantom{0} + \textsc{SeqLab} & 72.86          & 83.00          & 74.68          & 82.50          & 73.14                & 81.74          & 79.71          & 75.88          & 72.20          & 77.65          & 58.44          & 61.11          & 48.23          & 54.23          & 49.39                & 50.40          & 61.95          & 53.87          & 52.99          & 46.12          \\ \cdashlinelr{2-22}
                             & \textsc{ACS}                  & 76.14          & \textbf{87.58} & \textbf{78.94} & 83.84          & 75.51                & 81.35          & 81.33          & 74.21          & 73.73          & 77.30          & 60.58          & 63.41          & 49.99          & 56.38          & 52.98                & 51.72          & \textbf{66.58} & 55.90          & 53.57          & 46.07          \\
                             & \phantom{0} + \textsc{SeqLab} & \textbf{78.12} & 86.82          & 77.81          & \textbf{84.56} & \textbf{76.55}       & 81.36          & \textbf{82.10} & \textbf{76.51} & \textbf{74.74} & \textbf{79.85} & \textbf{62.85} & \textbf{63.47} & \textbf{52.01} & \textbf{58.48} & \textbf{53.39}       & 50.98          & 66.49          & \textbf{57.11} & \textbf{54.37} & \textbf{47.61} \\ \midrule
\multirow{10}{*}{\rotatebox[origin=c]{90}{restaurant}} & \textsc{Monolingual}          & 75.18          & 67.80          & 71.39          & 72.52          & 54.98                & --             & --             & --             & --             & --             & 71.20          & 58.93          & 64.35          & 69.57          & 36.84                & --             & --             & --             & --             & --             \\
                             & \phantom{0} + \textsc{SeqLab} & 77.18          & \textbf{70.35} & 73.84          & 73.23          & 62.55                & --             & --             & --             & --             & --             & 72.36          & 60.35          & 67.39          & 70.30          & 49.88                & --             & --             & --             & --             & --             \\ \cdashlinelr{2-22}
                             & \textsc{Multilingual}         & 79.06          & 65.64          & 78.63          & 77.31          & 56.82                & --             & --             & --             & --             & --             & 73.94          & 62.01          & 79.09          & 73.71          & 51.36                & --             & --             & --             & --             & --             \\
                             & \phantom{0} + \textsc{SeqLab} & \textbf{79.87} & 66.96          & \textbf{82.77} & \textbf{77.78} & \textbf{64.23}       & --             & --             & --             & --             & --             & \textbf{76.23} & \textbf{63.26} & \textbf{80.52} & \textbf{74.85} & \textbf{53.06}       & --             & --             & --             & --             & --             \\ \cdashlinelr{2-22}
                             & \textsc{Zero-shot}             & 62.30          & 57.83          & 66.46          & 60.64          & 46.23                & --             & --             & --             & --             & --             & 54.60          & 49.81          & 59.64          & 55.08          & 39.61                & --             & --             & --             & --             & --             \\
                             & \phantom{0} + \textsc{SeqLab} & 63.35          & 60.18          & 69.07          & 63.36          & 48.91                & --             & --             & --             & --             & --             & 58.00          & 51.79          & 62.25          & 58.05          & 43.42                & --             & --             & --             & --             & --             \\ \cdashlinelr{2-22}
                             & \textsc{Tr-only}              & 71.46          & 61.76          & 71.39          & 67.62          & 44.95 & --             & --             & --             & --             & --             & 65.21          & 54.21          & 65.89          & 63.88          & 46.22 & --             & --             & --             & --             & --             \\
                             & \phantom{0} + \textsc{SeqLab} & 72.21          & 62.45          & 72.27          & 69.30          & 46.72 & --             & --             & --             & --             & --             & 66.34          & 54.94          & 66.60          & 63.67          & 49.41                & --             & --             & --             & --             & --             \\ \cdashlinelr{2-22}
                             & \textsc{ACS}                  & 70.84          & 62.30          & 71.52          & 67.96          & 52.92                & --             & --             & --             & --             & --             & 64.98          & 55.14          & 65.79          & 63.55          & 49.41                & --             & --             & --             & --             & --             \\
                             & \phantom{0} + \textsc{SeqLab} & 72.56          & 62.42          & 72.00          & 68.99          & 53.02                & --             & --             & --             & --             & --             & 66.28          & 56.30          & 66.58          & 64.27          & 46.35                & --             & --             & --             & --             & --             \\ \bottomrule 
\end{tabular}
\end{adjustbox}
\caption{Complete results with different methods on different tasks and domains with English as the source language. \textbf{Bold} results are the best results overall for each domain, task, and language combination.}
\label{tab:res_all_en_source}
\end{table*}

Our prompt, inspired by \citet{smid-etal-2024-llama}, for GPT-4o mini defines the task, sentiment elements, and expected output. Figure~\ref{fig:prompt} shows the prompt for the TASD task, adaptable for E2E-ABSA by omitting the aspect category.

\begin{table*}[ht!]
    \centering
    \begin{adjustbox}{width=0.95\linewidth}
    \begin{tabular}{@{}clcccccccccccc@{}}
    \toprule
    \multirow{2}{*}{\textbf{\begin{tabular}[c]{@{}l@{}}Source\\ Language\end{tabular}}} & \multirow{2}{*}{\textbf{Method}} & \multicolumn{6}{c}{\textbf{E2E-ABSA}}         & \multicolumn{6}{c}{\textbf{TASD}}             \\ \cmidrule(lr){3-8} \cmidrule(lr){9-14}
                                                                                        &                                  & En    & Es    & Fr    & Nl    & Ru    & Tr    & En    & Es    & Fr    & Nl    & Ru    & Tr    \\ \midrule
    \multirow{3}{*}{Mono}                                                               & GPT-4o mini & 55.10 & 50.02 & 48.21 & 49.14 & 45.19 & 45.78 & 46.77 & 40.92 & 39.80 & 35.13 & 32.66 & 32.41 \\\cdashlinelr{2-14}
    
    & \textsc{Monolingual}              & 72.98 & 75.18 & 67.80 & 71.39 & 72.52 & 54.98 & 64.91 & 71.20 & 58.93 & 64.35 & 69.57 & 36.84 \\
                                                                                        & \phantom{0} + \textsc{SeqLab}     & 75.56 & 77.18 & 70.35 & 73.84 & 73.23 & 62.55 & 65.72 & 72.36 & 60.35 & 67.39 & 70.30 & 49.88 \\ \midrule
    \multirow{4}{*}{En}                                                                 & \textsc{Zero-shot}               & --     & 62.30 & 57.83 & 66.46 & 60.64 & 46.23 & --     & 54.60 & 49.81 & 59.64 & 55.08 & 39.61 \\
                                                                                        & \phantom{0} + \textsc{SeqLab}      & --     & 63.35 & 60.18 & 69.07 & 63.36 & 48.91 & --     & 58.00 & 51.79 & 62.25 & 58.05 & 43.42 \\ \cdashlinelr{2-14}
                                                                                        & \textsc{ACS}                     & --     & 70.84 & 62.30 & 71.52 & 67.96 & 52.92 & --     & 64.98 & 55.14 & 65.79 & 63.55 & 49.41 \\
                                                                                        & \phantom{0} + \textsc{SeqLab}            & --     & 72.56 & 62.42 & 72.00 & \textbf{68.99} & \textbf{53.02} & --     & 66.28 & 56.30 & \textbf{66.58} & \textbf{64.27} & 46.35 \\ \cdashlinelr{1-14}
    \multirow{4}{*}{Es}                                                                 & \textsc{Zero-shot}               & 61.02 & --     & 55.18 & 67.07 & 61.25 & 48.07 & 47.83 & --     & 48.38 & 60.44 & 58.29 & 42.10 \\
                                                                                        & \phantom{0} + \textsc{SeqLab}      & 64.24 & --     & 55.72 & 68.51 & 63.46 & 47.13 & 51.61 & --     & 48.87 & 61.61 & 57.24 & 43.94 \\ \cdashlinelr{2-14}
                                                                                        & \textsc{ACS}                     & 68.01 & --     & 63.68 & 71.20 & 68.00 & 52.05 & 62.76 & --     & 56.56 & 66.00 & 63.00 & 45.67 \\
                                                                                        & \phantom{0} + \textsc{SeqLab}            & \textbf{68.95} & --     & 63.82 & \textbf{72.50} & 68.49 & 51.51 & 63.22 & --     & 56.61 & 66.14 & 63.95 & \textbf{46.65} \\ \cdashlinelr{1-14}
    \multirow{4}{*}{Fr}                                                                 & \textsc{Zero-shot}               & 62.38 & 58.74 & --     & 65.35 & 56.35 & 43.24 & 53.78 & 52.42 & --     & 59.91 & 54.06 & 36.90 \\
                                                                                        & \phantom{0} + \textsc{SeqLab}      & 63.47 & 61.45 & --     & 67.57 & 59.99 & 46.19 & 53.87 & 56.35 & --     & 61.60 & 54.33 & 38.96 \\ \cdashlinelr{2-14}
                                                                                        & \textsc{ACS}                     & 67.78 & 73.86 & --     & 70.32 & 66.22 & 48.81 & 62.59 & 67.67 & --     & 64.64 & 62.93 & 43.88 \\
                                                                                        & \phantom{0} + \textsc{SeqLab}           & 68.32 & \textbf{74.28} & --     & 70.71 & 67.37 & 49.26 & \textbf{63.32} & \textbf{69.09} & --     & 64.86 & 63.04 & 43.40 \\ \cdashlinelr{1-14}
    \multirow{4}{*}{Nl}                                                                 & \textsc{Zero-shot}               & 61.06 & 58.35 & 57.66 & --     & 54.46 & 44.55 & 52.58 & 54.85 & 49.63 & --     & 51.44 & 36.66 \\
                                                                                        & \phantom{0} + \textsc{SeqLab}      & 64.06 & 62.14 & 59.62 & --     & 59.09 & 45.95 & 54.04 & 56.92 & 52.51 & --     & 56.49 & 40.35 \\ \cdashlinelr{2-14}
                                                                                        & \textsc{ACS}                     & 66.29 & 71.19 & 63.91 & --     & 66.72 & 45.09 & 61.55 & 65.97 & 57.79 & --     & 61.91 & 39.72 \\
                                                                                        & \phantom{0} + \textsc{SeqLab}            & 67.29 & 71.40 & \textbf{63.84} & --     & 66.97 & 46.58 & 60.80 & 66.21 & \textbf{58.03} & --     & 62.77 & 40.66 \\ \cdashlinelr{1-14}
    \multirow{4}{*}{Ru}                                                                 & \textsc{Zero-shot}               & 63.79 & 59.05 & 56.30 & 62.57 & --     & 50.07 & 46.04 & 50.70 & 44.63 & 58.02 & --     & 40.69 \\
                                                                                        & \phantom{0} + \textsc{SeqLab}      & 65.16 & 63.33 & 57.90 & 64.92 & --     & 50.38 & 48.02 & 54.93 & 48.73 & 59.58 & --     & 42.70 \\ \cdashlinelr{2-14}
                                                                                        & \textsc{ACS}                     & 57.81 & 60.98 & 53.29 & 65.15 & --     & 52.15 & 53.28 & 56.87 & 47.28 & 59.58 & --     & 46.84 \\
                                                                                        & \phantom{0} + \textsc{SeqLab}            & 57.66 & 62.10 & 52.46 & 64.46 & --     & 51.26 & 52.27 & 57.00 & 46.82 & 60.32 & --     & 45.44 \\ \cdashlinelr{1-14}
    \multirow{4}{*}{Tr}                                                                 & \textsc{Zero-shot}               & 44.53 & 39.94 & 28.77 & 43.19 & 31.81 & --     & 30.96 & 26.12 & 25.51 & 36.65 & 21.91 & --     \\
                                                                                        & \phantom{0} + \textsc{SeqLab}      & 47.98 & 37.23 & 35.18 & 46.57 & 39.68 & --     & 33.88 & 29.43 & 25.14 & 33.79 & 28.23 & --     \\ \cdashlinelr{2-14}
                                                                                        & \textsc{ACS}                     & 48.00 & 49.93 & 47.84 & 53.65 & 57.64 & --     & 43.17 & 46.15 & 41.10 & 48.97 & 52.49 & --     \\
                                                                                        & \phantom{0} + \textsc{SeqLab}            & 48.79 & 52.26 & 48.49 & 54.88 & 60.51 & --     & 44.37 & 48.22 & 41.92 & 49.06 & 55.32 & --    \\ \bottomrule
    \end{tabular}
\end{adjustbox}
\caption{Results on the E2E-ABSA and TASD tasks in the restaurant domain with different combinations of source and target languages. \textbf{Bold} results indicate the best outcomes for a given target language and task in cross-lingual settings.}
\label{tab:res_all}
\end{table*}

\section{Additional Results}
\label{app:results}

This section provides additional results on the restaurant domain, including results with different source--target language combinations and results with different backbone models, showcasing the effectiveness of the proposed approach, and provides detailed results for different domains with different methods.

\subsection{Detailed Domain Results}
\label{app:domain}

Table~\ref{tab:res_all_en_source} presents results across different training strategies. In addition to the \textsc{Monolingual}, \textsc{Zero-shot}, and \textsc{ACS} settings, we evaluate a multilingual setup, where models are trained on a combination of English and gold target-language data, and a \textsc{Tr-only} cross-lingual setting, which uses only translated data without aspect code-switching.

A clear pattern emerges across domains. In the restaurant domain, multilingual training consistently yields the best performance, closely followed by monolingual models, while cross-lingual approaches remain inferior. In particular, the \textsc{Tr-only} setting is noticeably weaker than multilingual training, highlighting the limitations of relying solely on translated data in this domain. In contrast, for the hotel and laptop domains, cross-lingual methods -- particularly \textsc{ACS + SeqLab} -- often outperform monolingual models, while multilingual results are generally comparable to, or slightly better than, monolingual performance.

We attribute this difference to the nature of the datasets. The restaurant domain consists of real-world, human-authored reviews, whereas the hotel and laptop datasets are parallel corpora obtained via translation from English. As a result, the gold target-language training data in these domains may be of lower quality or less diverse. At the same time, our \textsc{ACS} approach also relies on translating English data, producing training data that closely resembles the parallel datasets. This similarity is further amplified by the fact that the dataset creators~\citep{wu-etal-2025-absa} use the same translation engine and a similar aspect alignment strategy. Incorporating English data in the multilingual setting additionally introduces higher-quality supervision, which may explain the strong performance observed.

This hypothesis is supported by the near-identical performance of the multilingual and \textsc{Tr-only} settings in the hotel and laptop domains, suggesting a high degree of similarity between the translated and gold data. To quantify this, we measure similarity between the gold training data and our translated data using BERTScore~\citep{2020BERTScore} with XLM-R-large (token-level similarity) and SBERT~\citep{reimers-gurevych-2019-sentence} with multilingual E5~\citep{wang2024multilinguale5textembeddings} (sentence-level similarity). The average BERTScore reaches 99.87\%, and the SBERT score 99.28\%, indicating that the two datasets are nearly identical. This high overlap likely explains why translation-based approaches perform so similar to multilingual method in these domains.

\begin{table*}[ht!]
    \centering
    \begin{adjustbox}{width=0.99\linewidth}
    \begin{tabular}{@{}llcccccccccccc@{}}
    \toprule
    \multirow{2}{*}{\textbf{Model}} & \multirow{2}{*}{\textbf{Method}} & \multicolumn{6}{c}{\textbf{E2E-ABSA}}                                                               & \multicolumn{6}{c}{\textbf{TASD}}                                                                   \\ \cmidrule(lr){3-8} \cmidrule(lr){9-14}
                                    &                                  & Es             & Fr             & Nl             & Ru             & Tr             & Avg            & Es             & Fr             & Nl             & Ru             & Tr             & Avg            \\ \midrule
    \multirow{2}{*}{mT5 base}       & \textsc{Monolingual}             & 75.18          & 67.80          & 71.39          & 72.52          & 54.98          & 68.37          & 71.20          & 58.93          & 64.35          & 69.57          & 36.84          & 60.18          \\
                                    & \phantom{0} + \textsc{SeqLab}    & 77.18          & 70.35          & 73.84          & 73.23          & 62.55          & 71.43          & 72.36          & 60.35          & 67.39          & 70.30          & 49.88          & 64.06          \\ \cdashlinelr{1-14}
    \multirow{2}{*}{mT5 large}      & \textsc{Monolingual}             & 79.03          & 73.95          & 77.35          & 76.35          & 69.59          & 75.25          & 75.71          & 67.31          & 72.91          & 73.44          & 62.07          & 70.29          \\
                                    & \phantom{0} + \textsc{SeqLab}    & \textbf{79.81} & \textbf{74.80} & \textbf{77.77} & \textbf{77.08} & \textbf{70.65} & \textbf{76.02} & \textbf{76.31} & \textbf{68.27} & \textbf{73.68} & \textbf{74.11} & \textbf{64.73} & \textbf{71.42} \\ \cdashlinelr{1-14}
    \multirow{2}{*}{mBART large}    & \textsc{Monolingual}             & 77.26          & 71.54          & 77.05          & 74.08          & 68.87          & 73.76          & 73.25          & 65.27          & 71.55          & 70.93          & 58.60          & 67.92          \\
                                    & \phantom{0} + \textsc{SeqLab}    & 77.45          & 72.75          & 77.20          & 74.50          & 68.59          & 74.10          & 74.04          & 65.93          & 72.86          & 71.62          & 59.73          & 68.84          \\ \midrule
    \multirow{4}{*}{mT5 base}       & \textsc{Zero-shot}               & 62.30          & 57.83          & 66.46          & 60.64          & 46.23          & 58.69          & 54.60          & 49.81          & 59.64          & 55.08          & 39.61          & 51.75          \\
                                    & \phantom{0} + \textsc{SeqLab}      & 63.35          & 60.18          & 69.07          & 63.36          & 48.91          & 60.97          & 58.00          & 51.79          & 62.25          & 58.05          & 43.42          & 54.70          \\ \cdashlinelr{2-14}
                                    & \textsc{ACS}                     & 70.84          & 62.30          & 71.52          & 67.96          & 52.92          & 65.11          & 64.98          & 55.14          & 65.79          & 63.55          & 49.41          & 59.77          \\
                                    & \phantom{0} + \textsc{SeqLab}            & 72.56          & 62.42          & 72.00          & 68.99          & 53.02          & 65.80          & 66.28          & 56.30          & 66.58          & 64.27          & 46.35          & 59.95          \\ \cdashlinelr{1-14}
    \multirow{4}{*}{mT5 large}      & \textsc{Zero-shot}               & 67.32          & 65.45          & 72.82          & 67.56          & 56.43          & 65.92          & 61.42          & 58.54          & 66.73          & 64.36          & 51.08          & 60.43          \\
                                    & \phantom{0} + \textsc{SeqLab}      & 70.27          & 66.85          & 73.87          & 68.22          & 56.54          & 67.15          & 64.63          & 60.15          & 68.04          & 64.66          & 52.27          & 61.95          \\ \cdashlinelr{2-14}
                                    & \textsc{ACS}                     & 75.09          & 66.91          & 74.48          & 71.04          & 58.17          & 69.14          & 69.80          & 59.85          & 68.47          & 67.54          & 50.91          & 63.31          \\
                                    & \phantom{0} + \textsc{SeqLab}            & \textbf{75.97} & \textbf{67.70} & \textbf{75.80} & \textbf{72.48} & \textbf{59.23} & \textbf{70.23} & \textbf{70.99} & \textbf{60.97} & \textbf{69.74} & \textbf{68.27} & \textbf{54.02} & \textbf{64.80} \\ \cdashlinelr{1-14}
    \multirow{4}{*}{mBART large}    & \textsc{Zero-shot}               & 65.44          & 60.95          & 70.31          & 66.55          & 40.66          & 60.78          & 59.44          & 55.31          & 63.23          & 61.35          & 36.34          & 55.13          \\
                                    & \phantom{0} + \textsc{SeqLab}      & 67.58          & 61.08          & 71.71          & 65.32          & 40.18          & 61.17          & 62.21          & 55.93          & 64.80          & 61.00          & 38.90          & 56.57          \\ \cdashlinelr{2-14}
                                    & \textsc{ACS}                     & 73.46          & 62.73          & 73.42          & 69.83          & 55.70          & 67.03          & 67.54          & 57.35          & 68.28          & 65.23          & 48.52          & 61.38          \\
                                    & \phantom{0} + \textsc{SeqLab}            & 74.09          & 63.31          & 73.98          & 70.40          & 56.78          & 67.71          & 68.74          & 57.87          & 67.73          & 66.22          & 50.06          & 62.12          \\ \bottomrule
    \end{tabular}
\end{adjustbox}
\caption{Results with different models in the restaurant domain. \textbf{Bold} results indicate the best performance for each task and dataset in both monolingual and cross-lingual settings.}
\label{tab:res_models}
\end{table*}

Overall, the best performance with high-quality target-language data is achieved through multilingual or monolingual training, while \textsc{ACS + SeqLab} proves most effective when such data are unavailable or of lower quality. The \textsc{ACS} approach consistently outperforms the \textsc{Tr-only} setting, whereas \textsc{Zero-shot} remains the weakest strategy. Additionally, incorporating \textsc{SeqLab} generally leads to further performance improvements across settings.

\subsection{Different Source--Target Combinations}
Table~\ref{tab:res_all} presents results for both E2E-ABSA and TASD across various source-target language combinations. The trends largely align with the previous findings. The \textsc{SeqLab} method consistently improves results across languages and tasks. Combining \textsc{SeqLab} with \textsc{ACS} data further enhances performance. TASD is more challenging than E2E-ABSA, as it requires predicting sentiment element triplets rather than simple tuples.

A notable observation is that the optimal source language varies for different target languages. For instance, Spanish-French transfer performs exceptionally well, likely due to the shared linguistic characteristics of these Romance languages. In contrast, Turkish consistently yields the lowest performance as a source and target language, likely due to its structural differences from the other evaluated languages.

Our approach also clearly outperforms GPT-4o mini, especially on TASD, with gains of over 20\% in most cross-lingual settings compared to its zero-shot results.

Overall, the combination of \textsc{ACS} and \textsc{SeqLab} is a robust strategy across multiple tasks and language pairs for cross-lingual ABSA.

\subsection{Models Comparison}
Table~\ref{tab:res_models} compares performance across three different models on E2E-ABSA and TASD with English as the source language. As expected, larger models yield better results, with mT5 large achieving the best overall performance. The \textsc{ACS} and \textsc{SeqLab} methods significantly boost performance across all models. The best results for each model are obtained when combining \textsc{ACS} and \textsc{SeqLab}, demonstrating the versatility and effectiveness of these methods across different architectures, specifically mT5 and mBART.

\begin{table*}[ht!]
\centering
\begin{adjustbox}{width=0.99\linewidth}
\begin{tabular}{@{}lllll@{}}
\toprule
Language & Sentence & Gold Aspect & Predicted Aspect & Error Type \\ \midrule
Spanish & \begin{tabular}[c]{@{}l@{}}Buena localización en el centro de la Barcelona delant de mar .\end{tabular} & localización & – & Missed aspect \\ \cdashlinelr{1-5}
Spanish & \begin{tabular}[c]{@{}l@{}}Solamente el vino rosado es un poquito mediano\end{tabular} & vino & vino rosado & Over-predicted aspect \\\cdashlinelr{1-5}

French & \begin{tabular}[c]{@{}l@{}}Ne surtout pas y aller pour bien manger ,\\ par contre on se laisse bien embarquer dans l' ambiance de ces années là .\end{tabular} & ambiance & – & Missed aspect \\\cdashlinelr{1-5}
French & \begin{tabular}[c]{@{}l@{}}Une adresse à conseiller si vous aimez la cuisine traditionnelle ,\\ avec juste ce qu' il faut de goût .\end{tabular} & cuisine traditionnelle & cuisine & Partial aspect extraction \\\cdashlinelr{1-5}

Dutch & \begin{tabular}[c]{@{}l@{}}Mijn tweede Leefe uit het vatkwam in een flesje !\end{tabular} & Leefe & – & Missed aspect \\\cdashlinelr{1-5}
Dutch & \begin{tabular}[c]{@{}l@{}}Tafel gereserveerd en dan genieten maar , dachten we .\end{tabular} & – & Tafel & False positive \\\cdashlinelr{1-5}

Russian & \begin{tabular}[c]{@{}l@{}}Est' zal dlja nekurjaŝih , čto nesomnenno važno .\end{tabular} & zal dlja nekurjaŝih & zal & Partial aspect extraction \\\cdashlinelr{1-5}
Russian & \begin{tabular}[c]{@{}l@{}}Odnako , svobodnyh stolikov bylo mnogo .\end{tabular} & – & stolikov & False positive \\\cdashlinelr{1-5}

Turkish & \begin{tabular}[c]{@{}l@{}}Istiklal Caddesi ' ne girdiğinizde solda kalıyor .\end{tabular} & solda & Istiklal Caddesi & Wrong aspect extraction \\\cdashlinelr{1-5}
Turkish & \begin{tabular}[c]{@{}l@{}}Lahmacun ve içliköfte iyi fakat dolmaları Antep\\ usulünde İstanbul'da yediklerim arasında en iyilerden di.\end{tabular} & Lahmacun ve içliköfte & Lahmacun & Partial aspect extraction \\
 \bottomrule
\end{tabular}
\end{adjustbox}
\caption{Illustrative examples of errors made by the \textsc{ACS + SeqLab} approach using the mT5-base model on the restaurant dataset. Cyrillic text has been transliterated to Latin here for readability; the model uses the original script as input.}
\label{tab:qualitative}
\end{table*}

\section{Qualitative Error Analysis}
\label{app:qualitative}
Table~\ref{tab:qualitative} presents a few representative examples of model errors on the restaurant dataset with the \textsc{ACS + SeqLab} approach. 
We categorize these errors into four types for clarity: 
\begin{itemize}
    \item \textbf{Partial aspect extraction:} The model predicts only part of a multi-word aspect.
    \item \textbf{Over-predicted aspect:} The model predicts extra tokens beyond the gold aspect.
    \item \textbf{Wrong aspect extraction:} The model predicts an unrelated aspect.
    \item \textbf{False positive:} The model predicts an aspect when none is present.
\end{itemize}

\end{document}